\documentclass{article}
\usepackage{PRIMEarxiv}
\usepackage[normalem]{ulem} 
\usepackage[utf8]{inputenc} 
\usepackage[T1]{fontenc}    
\usepackage{hyperref}       
\usepackage{url}            
\usepackage{booktabs}       
\usepackage{amsfonts}       
\usepackage{nicefrac}       
\usepackage{microtype}      
\usepackage{lipsum}
\usepackage{fancyhdr}       
\usepackage{graphicx}       
\graphicspath{{media/}}     
\usepackage{xcolor}

\usepackage{multicol}
\usepackage{amsmath}
\usepackage{multirow}

\usepackage{graphicx}
\usepackage{caption}
\usepackage{subcaption}

\usepackage{soul}

\renewcommand{\arraystretch}{1.1}
  
\title{CHaRM: Conditioned Heatmap Regression Methodology for Accurate and Fast Dental Landmark Localization
}

\author{
    José Rodríguez-Ortega \\
    Nemotec \\
    Madrid, Spain \\
    \texttt{joserodriguez@nemotec.com} \\
    \\
    Dept. of Computer Science and Artificial Intelligence \\ 
    University of Granada\\
    Granada, Spain \\
    \texttt{e.jrodriguez98@go.ugr.es} \\
    \And
    Francisco Pérez-Hernández \\
    Nemotec \\
    Madrid, Spain \\
    \texttt{franciscoperez@nemotec.com} \\
   \And
    Siham Tabik \\
    Dept. of Computer Science and Artificial Intelligence, \\
    University of Granada\\
    Granada, Spain \\
    \texttt{siham@ugr.es} \\
}

\begin{document}
\maketitle

\begin{abstract}
Identifying anatomical landmarks in 3D dental models is essential for orthodontic treatment, yet manual placement is labor-intensive and requires expert knowledge. While machine learning methods have been proposed for automatic landmark detection in 3D Intraoral Scans (IOS), none provide a fully end-to-end solution that avoids costly tooth segmentation.

We present CHaRM (Conditioned Heatmap Regression Methodology), the first fully end-to-end deep learning approach for tooth landmark detection in 3D IOS. CHaRM integrates four components: a point cloud encoder, a decoder with a heatmap regression head, a teeth-presence classification head, and the novel CHaR module. The CHaR module leverages teeth-presence information to adapt to missing teeth, improving detection accuracy in complex dental cases. Unlike two-stage workflows that segment teeth before landmarking, CHaRM operates directly on IOS point clouds, reducing complexity, avoiding error propagation, and lowering computational cost.

We evaluated CHaRM with five point cloud learning backbones on IOSLandmarks-1k, a new dataset of 1,214 annotated 3D dental models. Both the dataset and code will be publicly released\footnote{Resources will be made available upon acceptance} to address the scarcity of open data in orthodontics and foster reproducible research.

CHaRM with PointMLP, named CHaRNet, achieved the best accuracy and efficiency. Compared to state-of-the-art methods (TSMDL and ALIIOS), CHaRNet reduced mean Euclidean distance error to 0.56 mm on standard dental models and 1.12 mm across all dentition type, while delivering up to 14.8× faster inference on GPU.
This end-to-end approach streamlines orthodontic workflows, enhances the precision of 3D IOS analysis, and enables efficient computer-assisted treatment planning.

\end{abstract}


\section{Introduction}
Automatic 3D anatomical keypoints or landmarks detection plays a crucial role in computer-aided orthodontics, allowing tasks such as orthodontic planning, prosthetic design, and the diagnosis of dental anomalies \cite{kumar2012automatic}. Intraoral scans (IOS) are commonly utilized to capture accurate digital surface models of dentitions. These scanners represent the 3D surface of teeth, typically as point clouds or meshes. These digital representations are invaluable for simulating procedures such as tooth extraction, movement, deletion, and rearrangement, allowing dentists to predict treatment outcomes more effectively. Consequently, digital dental models have the potential to streamline dentists' workflows and reduce the time spent on labor-intensive tasks.

Existing approaches to perform 3D dental tasks such as teeth segmentation and landmark detection often rely on point cloud-based learning methods \cite{tian2019automatic, wei2022dense, im2022accuracy, lian2020deep, lian2019meshsnet}. Although effective for general point cloud problems, these methods often struggle in dental applications due to the specialized nature of dental anatomies, which frequently exhibit incomplete geometries and subtle variations, such as missing teeth or the presence of third molars \cite{sun2020automatic, wei2022dense, wu2022two, baquero2022automatic}.    Moreover, traditional landmark detection workflows typically involve a preliminary tooth segmentation stage \cite{wu2022two, baquero2022automatic}. While widely adopted in digital dentistry, teeth segmentation introduces significant drawbacks. It is computationally demanding, requires careful post-processing to ensure stable contours, and most critically, can propagate segmentation errors into subsequent landmark detection methods, potentially compromising accuracy. In many clinical scenarios, such as prosthetic design or simple orthodontic assessments, explicit tooth segmentation may not be clinically necessary, yet current pipelines mandate its inclusion, adding unnecessary overhead.

Another significant limitation in this field is the scarcity of publicly available dental datasets. This restricts progress by preventing fair comparisons among methods and limiting accessibility for researchers from outside specialized dental domains \cite{wu2022two, baquero2022automatic}.

In this work, we address these challenges through three key contributions: 
\begin{enumerate}
    \item  We propose the Conditioned Heatmap Regression (CHaR) module,  specifically designed for point cloud learning architectures. This module dynamically adapts landmark detection to account for missing teeth, significantly improving accuracy across complex dental anatomies.
    \item We introduce CHaRM, an end-to-end Deep Learning (DL) methodology that directly detects landmarks on the IOS point cloud without relying on previous teeth segmentation, drastically reducing the latency of the overall computation. Concretely, CHaRNet, CHaRM based on PointMLP as point cloud processing model, is more accurate and several orders of magnitude faster than the state-of-the art (SOTA) TSMDL \cite{wu2022two} and ALIIOS \cite{baquero2022automatic}, in both CPU and GPU.
    \item We construct and make publicly available IOSLandmarks-1k, a dataset of $1,214$ 3D digital teeth models with tooth segmentation and landmarks annotations, with the hope to foster broader research and benchmarking. IOSLandmarks-1k includes diverse complexities and a taxonomy of dentition types to evaluate models across varying levels of complexity.
    
\end{enumerate}

Our results highlight the effectiveness of the proposed CHaRM and CHaR module, demonstrating significant performance improvements across multiple point cloud learning models. Notably, CHarNet achieves a Mean Euclidean Distance Error (MEDE) of 0.56 mm on standard dental models, defined as those with a complete set of teeth excluding third molars, and 1.12 mm macro-averaging across all dentition types and a Mean Success Ratio (MSR) of 85.2\% and 68.5\%, respectively, surpassing previous SOTA methods. Furthermore, by eliminating the need for previous teeth segmentation, CHaRNet speeds up inference time several orders of magnitude. These findings underscore the robustness and efficiency of CHaRNet, particularly in challenging cases involving missing teeth and/or third molars.

The remainder of this paper is organized as follows: Section \ref{sec:related} presents related works. Section \ref{sec:data} describes the dataset construction process, including the proposed taxonomy and data preprocessing steps. Section \ref{sec:methods} formally introduces the problem and the proposed CHaRM, detailing the CHaR module and its integration with point cloud learning architectures. Section \ref{sec:experiments} presents a thorough evaluation of the proposed approach across different point cloud architectures and the comparison with SOTA methods. Section \ref{sec:discussion} discusses the broader implications, limitations, and potential directions for future work. Finally, Section \ref{sec:conclusion} summarizes the key contributions and results of this work.

\section{Related works} \label{sec:related}

This section reviews related work in two areas: point cloud learning methods (Section \ref{sec:pcl}) and landmark detection approaches (Section \ref{sec:ldm}).
\subsection{Point cloud learning} \label{sec:pcl}
 At present, three main stream methods have been used for 3D point cloud processing: projection-based, voxel-based and point cloud-based. Projection-based methods try to firstly project the original point cloud to a simpler 2D domain to later apply common 2D operations for feature extraction \cite{chen2017multi, kanezaki2018rotationnet, lang2019pointpillars, su2015multi}. While simple and effective, the projection operations over point sets inherently collapse potentially useful geometric information. Voxel-based methods transform the irregular nature of point clouds to regular 3D voxels (the equivalent of pixels in 3D) then apply 3D convolution operations \cite{maturana2015voxnet, song2017semantic}. Given the sparse nature of 3D data, directly applying 3D convolutions over the voxel input results in a highly inefficient computation. To take advantage of this sparsity, sparse convolutions \cite{choy20194d, graham20183d, wang2017cnn} greatly reduce computation and memory requirements by evaluating the convolution operation only in occupied voxels. Finally, point cloud-based methods directly process point clouds as unordered point sets. PointNet \cite{qi2017pointnet} pioneered this approach by directly processing point sets using shared multilayer perceptrons (MLP). PointNet++ \cite{qi2017pointnet++} improves its predecessor by hierarchically applying  set abstraction operations. PointNeXt \cite{qian2022pointnext} further improves PointNet++ by using modern training strategies such as data augmentation, new optimization techniques, and increased model size. The authors of PointMLP \cite{ma2022rethinking} hypothesize that local geometrical information may not be the key to point cloud learning and introduce a simple and pure residual MLP network that performs competitively with more complicated methods. DGCNN \cite{wang2019dynamic} and similar graph-based approaches \cite{simonovsky2017dynamic, zhao2019pointweb, li2019deepgcns} perform typical graph operations like message passing over the point cloud previously connected as some form of graph. Given the success of the self-attention mechanism, PointTransformer \cite{zhao2021point} designed self-attention layers for point cloud processing and use them to perform tasks such as scene semantic segmentation or object classification in 3D scenes. Since then several studies have kept using  transformers \cite{guo2021pct, yang2023swin3d, robert2023efficient, wu2022point, wu2024point}. 

Recently, some DL methods  operate directly on the 3D point cloud representation of dental models.  For example,  \cite{zanjani2019deep} proposed an end-to-end DL approach based on PointCNN \cite{li2018pointcnn} to segment individual teeth and gingiva from point cloud representation of IOS. \cite{sun2020tooth} made use of FeaStNet \cite{verma2018feastnet}, a graph convolutional neural network, to also perform teeth segmentation on 3D dental models. The authors in \cite{zanjani2019mask} introduced Mask-MCNet to perform instance segmentation of point cloud data from IOS. Mask-MCNet localizes each tooth by predicting its 3D bounding box and simultaneously segments all the points inside each box. TSegNet \cite{cui2021tsegnet} was proposed as an end-to-end learning-based method for robust tooth segmentation on 3D point cloud dental models. It introduced a distance-aware tooth centroid voting scheme and a confidence-aware cascade segmentation module to handle challenging cases like missing, crowded, or misaligned teeth. DBGANet \cite{lin2023dbganet} was proposed as a Dual-Branch Geometric Attention Network for 3D tooth segmentation, utilizing centroid-guided separable attention and Gaussian neighbor attention to capture global geometric structure and refine tooth-gingiva boundaries. DentalMAE \cite{almalki2024self} extended the self-supervised MeshMAE \cite{liang2022meshmae} framework, demonstrating improved generalization and transfer learning for teeth segmentation on 3D intra-oral scans, even with limited training data.

\subsection{Landmark detection} \label{sec:ldm}
Landmark detection is a fundamental task in both computer vision and medical imaging, playing a critical role in tasks such as diagnosis, treatment planning, and surgical simulation. Most classical methods \cite{maes2010feature, sipiran2010robust, godil2011salient, li2013curvature} have historically made use of local geometric information to detect landmarks that are highly associated with sharp features. Although these methods are effective within their well-defined domains, they are limited in scope and are unable to identify landmarks outside their domain. With the rise of DL, two approaches dominate the data-driven landmark localization problem. The most intuitive approximation is the regression-based approach, which directly regresses coordinates from the input (images, volumes, point clouds, etc.) \cite{toshev2014deeppose, carreira2016human, xu2022vitpose, wang2024spatial}. However, this direct mapping is extremely challenging given the unbounded nature of the coordinates \cite{pfister2015flowing}. Instead, heatmap-based detection, introduced in \cite{tompson2014joint}, has become a popular and preferred approach for landmark detection due to its ability to encode location information as probabilities over the inputs. This approach simplifies the task by reducing the space of possibilities to the number of elements in the input space (e.g., pixels, voxels, vertices) \cite{shu2018detecting, sun2019deep, yang2021transpose}.

In recent years, landmark localization task has become popular in medical image analysis \cite{zhang2016detecting, lian2018hierarchical, zhang2020context, gong2024nnmamba}. However, only a handful of works have been proposed for landmark detection in 3D dental models. \cite{kumar2012automatic} proposed a set of analytical approaches to identify dental-specific features (e.g., cusps, marginal ridges, and grooves) on digital dental meshes. At the individual tooth level, \cite{wei2022dense} designed a novel neural network architecture for the joint tasks of predicting landmarks and axes. \cite{kubik2022robust} presented a method for landmark detection in 3D dental meshes that leverages a multi-view approach, transforming the task into the 2D domain. In this way, the network detects landmarks through heatmap regression across multiple viewpoints. ALR \cite{woodsend2022development} first determines the orientation of the scan, then uses local maxima along the vertical axis as an initial approximation for landmarks. This is followed by analyzing surface gradient and curvature information to identify the shape and boundaries of each tooth. c-SCN \cite{sun2020automatic} was proposed as an end-to-end method for tooth segmentation and landmark localization but only on teeth crowns. TS-MDL \cite{wu2022two} proposed a two-stage framework to subsequently perform teeth segmentation and landmark detection on IOS. By harnessing the segmentation outputs, authors used a variant of PointNet to detect landmarks for each individual tooth. While effective, this approach comes at a high computational cost derived from separating the process into two stages. Besides, the dataset used includes mainly regular dental models without considering  complex dental models with missing teeth and one or two third molars. ALIIOS algorithm \cite{baquero2022automatic} combines image processing, segmentation, and DL to identify dental landmarks on IOS by synthesizing 3D patches with the output of a 2D U-Net. However, most of these methods rely on a previous segmentation of teeth, which adds considerable latency to the overall system.

\renewcommand{\arraystretch}{1.3}
\begin{table}[t]
\centering
\caption{Comparison between datasets for landmark detection in 3D dental models.}
\resizebox{\columnwidth}{!}{%
\begin{tabular}{lllllr}
\hline
Work (year) & Missing teeth & Third molars & Dental arch & Public & \# of 3D dental models \\
 \hline
\cite{sun2020automatic} (2020) & Not mentioned & Not mentioned  & Not mentioned & No & $100$ \\

\cite{baquero2022automatic} (2022) & Yes & Yes & Upper and lower & No & $405$ \\

\cite{wu2022two} (2022) & Yes & No & Upper & No & $136$ \\

\cite{kubik2022robust} (2022) & Yes & Yes & Upper and lower & No & $337$ \\

Teeth3DS+ \cite{ben2022teeth3ds+} (2024) & Yes & Yes & Upper and lower & Yes & $340$ \\

IOSLandmarks-1k (2025) & Yes & Yes & Upper and lower & Yes & $1,214$ \\
\hline
\end{tabular}
}
\label{table:datasets}
\end{table}

\begin{figure}[t]
\begin{subfigure}{.48\textwidth}
  \centering
  \includegraphics[width=.8\linewidth]{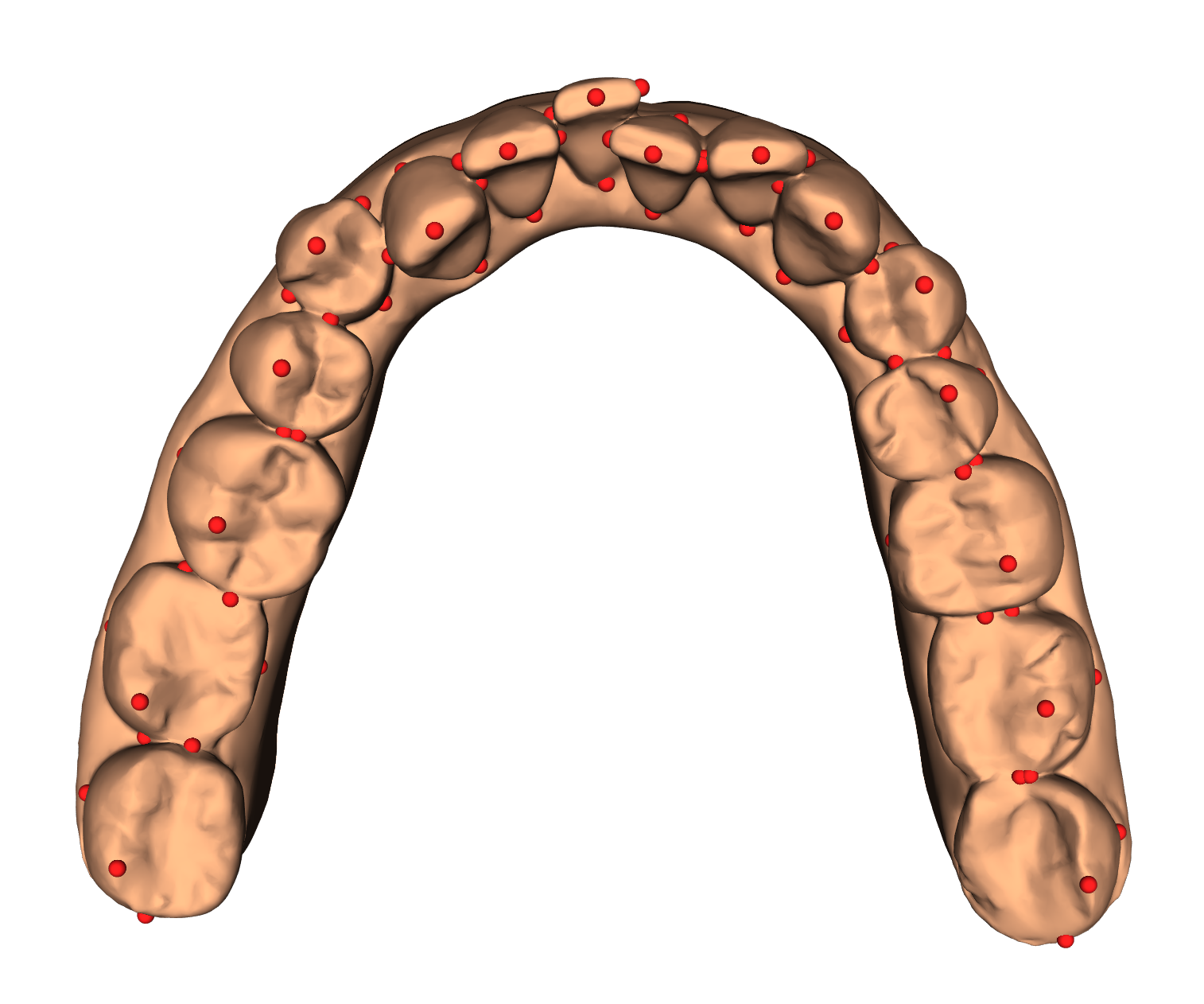}
  \caption{Occlusal view.}
  \label{fig:occlusal}
\end{subfigure}%
\begin{subfigure}{.48\textwidth}
  \centering
  \includegraphics[width=.8\linewidth]{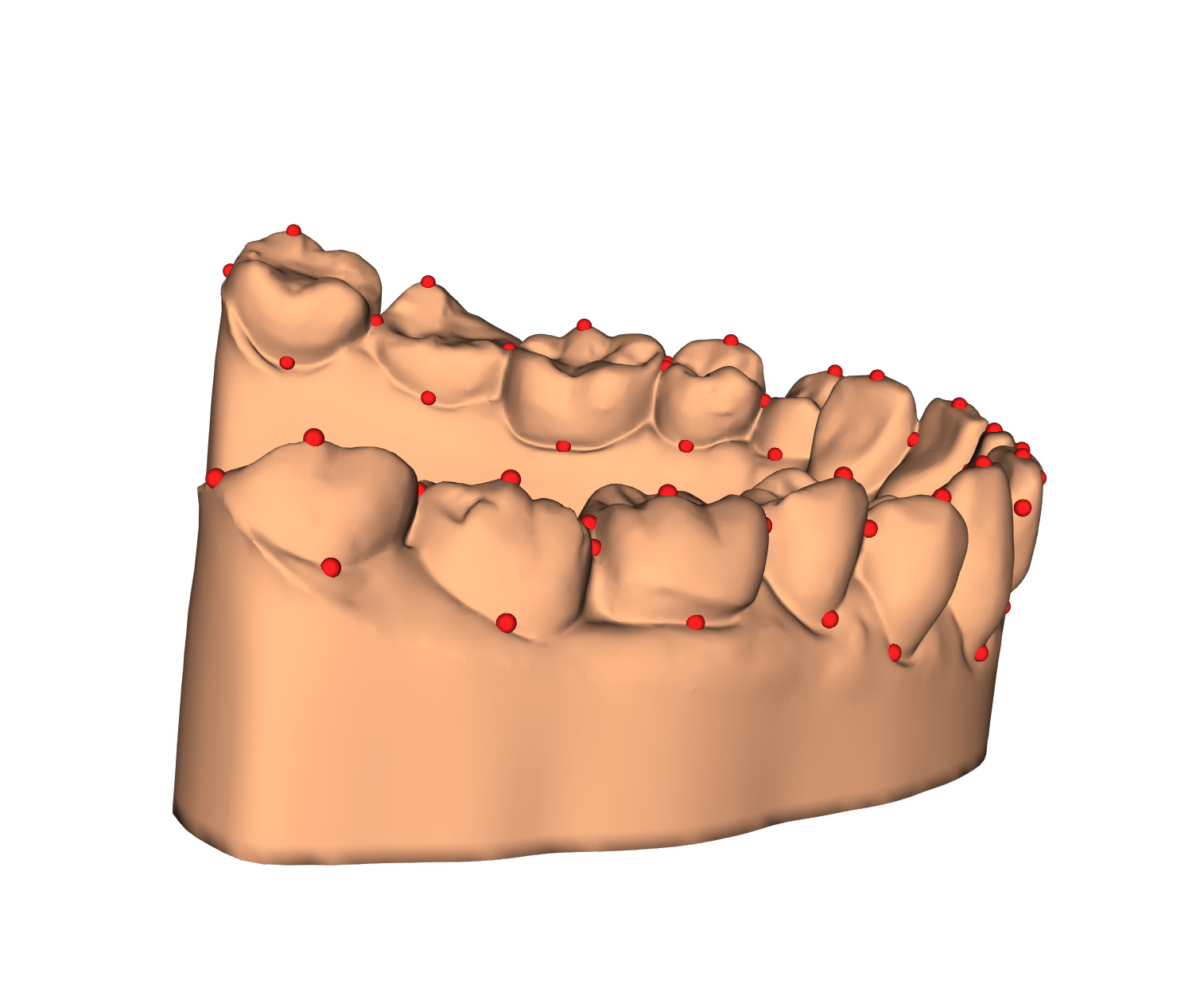}
  \caption{Lateral view.}
  \label{fig:lateral}
\end{subfigure}
\begin{subfigure}{.48\textwidth}
  \centering
  \includegraphics[width=.8\linewidth]{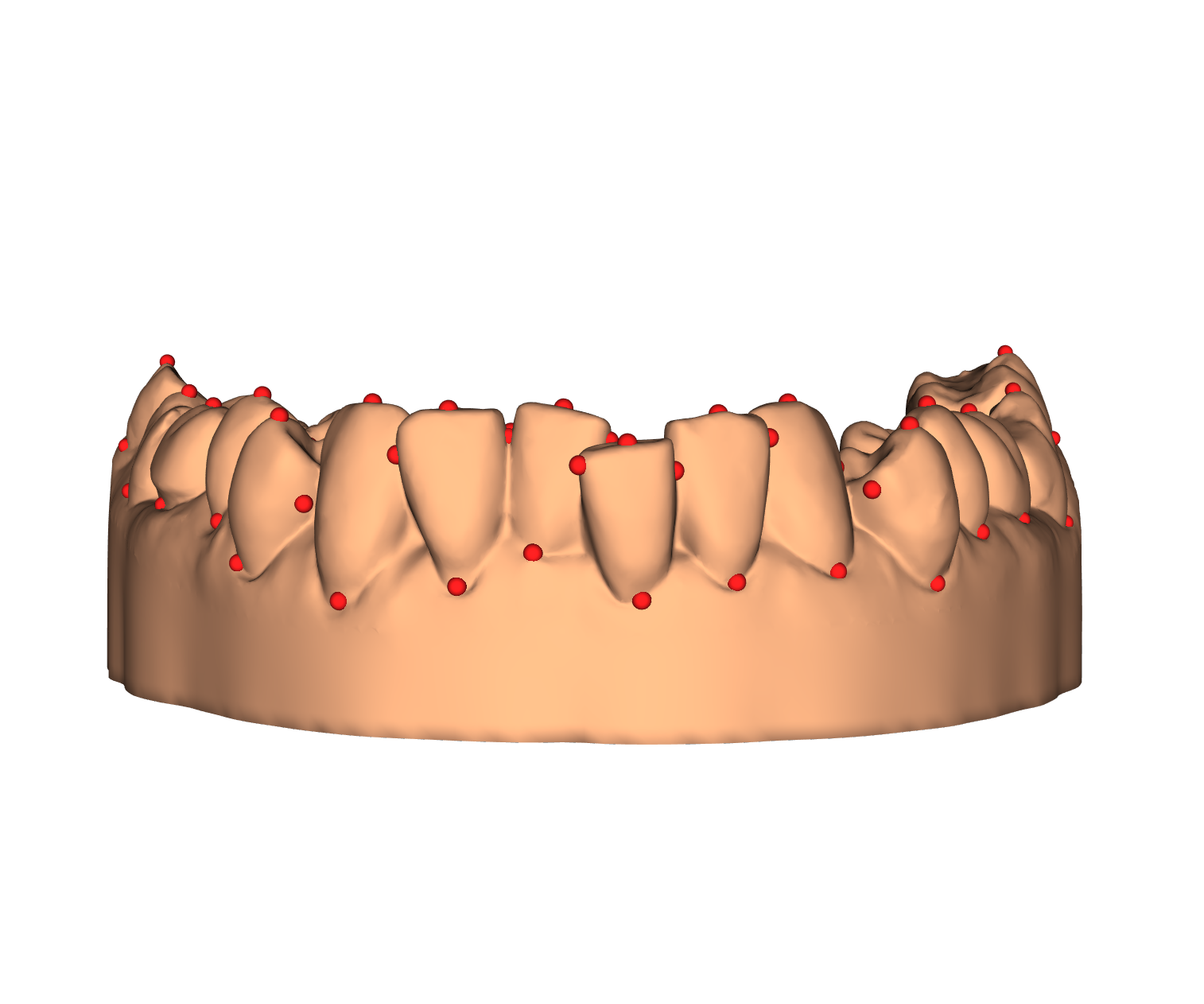}
  \caption{Facial view.}
  \label{fig:facial}
\end{subfigure}
\begin{subfigure}{.48\textwidth}
  \centering
  \includegraphics[width=.8\linewidth]{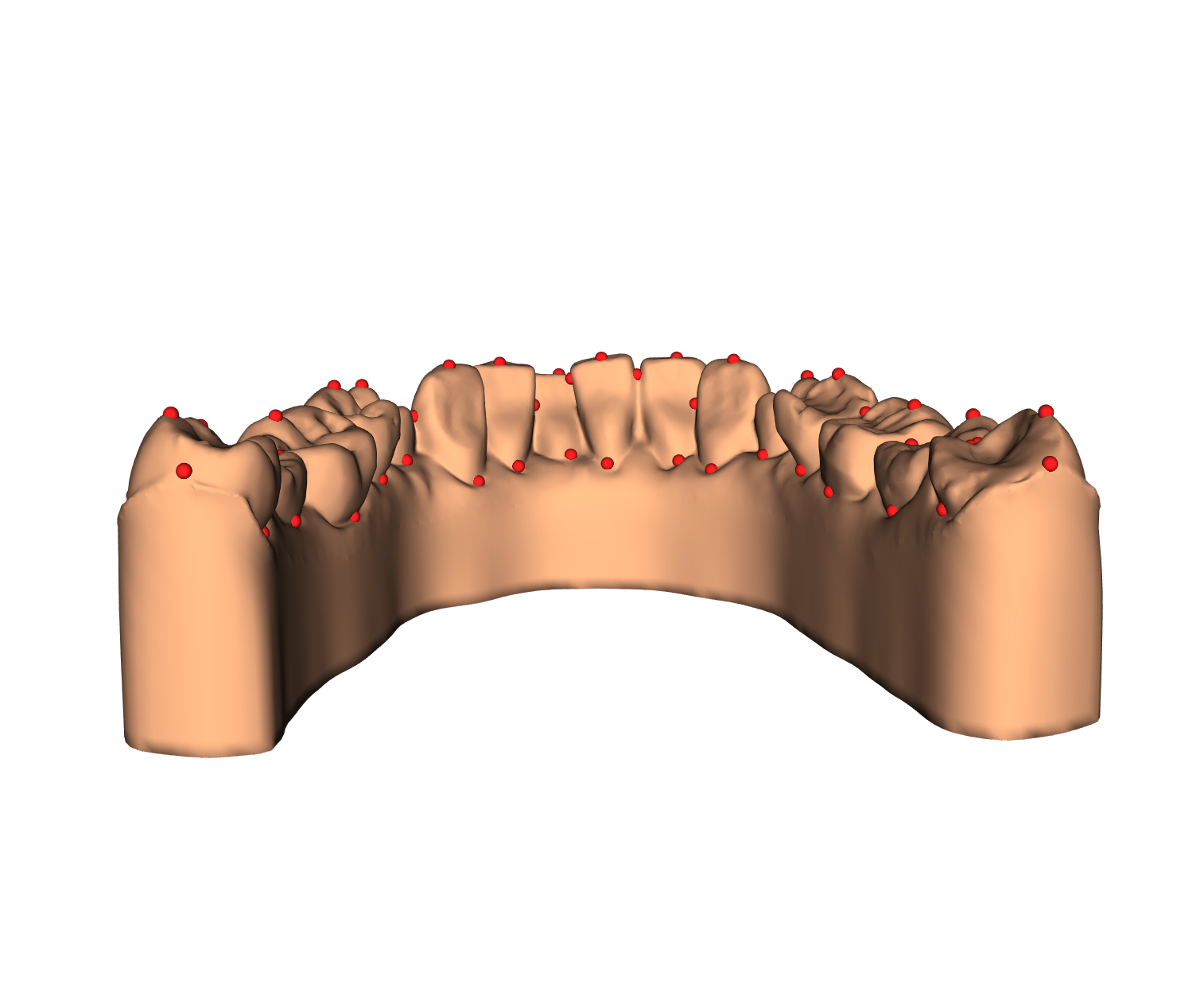}
  \caption{Lingual view.}
  \label{fig:lingual}
\end{subfigure}
\caption{Four views of a complete dentition (16 teeth) with its corresponding 80 landmarks (5 per tooth).}
\label{fig:views}
\end{figure}

\begin{table}[t]
\centering
\caption{Description of dental landmarks.}
\resizebox{\columnwidth}{!}{%
\begin{tabular}{llp{10cm}}
\hline
Landmark & Abbreviation & Description \\
 \hline
Mesial Point & MP & Located on the mesial surface of the tooth, facing towards the midline of the dental arch, indicating the most anterior point. \\

Distal Point & DP & Located on the distal surface of the tooth, facing away from the midline, indicating the most posterior point. \\

Cusp Point & CP & The tip of a cusp, typically found on the chewing surface of teeth, marking the highest point. \\

Facial Gingival Point & FGP & Located on the facial (or buccal) surface near the gingival margin on the cheek or lip side, indicating the gum line position. \\

Lingual Gingival Point & LGP & Located on the lingual (or palatal) surface near the gingival margin on the side facing the tongue, indicating the gum line position. \\
\hline
\end{tabular}
}
\label{table:dental_landmarks}
\end{table}

\section{IOSLandmarks-1k} \label{sec:data}

\begin{figure*}[t]
    \centering
    \includegraphics[width=1\textwidth]{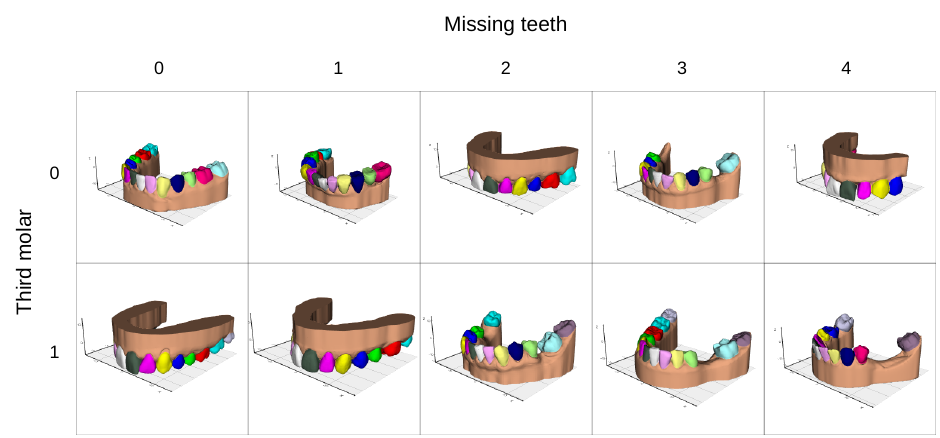}
    \caption{Examples of each dentition type based on our proposed taxonomy. The rows represent the presence or absence of third molars: the first row (0) indicates no third molars are present, while the second row (1) indicates the presence of at least one third molar. The columns correspond to the number of missing teeth in the dentition.  Each tooth is colored with its associated color given a predefined color map.}\
    \label{fig:examples}
\end{figure*}

Existing datasets for landmark detection in 3D intraoral scans are very small, private and/or do not include frequent anomalies of the dentition. A comparison between these datasets is provided in Table~\ref{table:datasets}. Teeth3DS+  \cite{ben2022teeth3ds+} is the only publicly available dataset. However, it only contains 340 annotated samples, which may be insufficient for training modern neural networks.  In this work, we construct IOSLandmarks-1k, a comprehensive dataset of 1,214 digital 3D dental models derived from post-processed raw scans. The post-processing step involves removing common artifacts found in raw scans, resulting in standardized dental models that retain only clinically relevant information. Each model is annotated with detailed reference landmarks and categorized into a taxonomy of dentition types to facilitate the evaluation of automatic 3D landmark detection systems under varying levels of difficulty. Furthermore, teeth segmentation annotations are also provided in order to also train and evaluate methods that require previous segmentation like TSMDL and ALIIOS (see Figure \ref{fig:examples}). IOSLandmarks-1k does not only provide a robust foundation for training and evaluating teeth segmentation and landmark detection methods, but also addresses a significant gap in the orthodontic field, where publicly available annotated datasets are scarce. In this regard, we make IOSLandmarks-1k publicly available so that the community can explore new research ideas on it. This is also an attempt to encourage the orthodontic community to publish their datasets and hence allow for a fair comparison between different AI methods, attracting more practitioners to the field of computer-assisted orthodontics and encouraging the development of new domain-specific technologies. Full details on the segmentation and landmark annotations format will be provided in  OSLandmarks-1k  public repository.

To enable accurate and consistent annotations, each dentition in our dataset contains five landmarks per tooth, making up to 80 landmarks in total (depending on the number of teeth present in each dentition). A visual  illustration is provided in Figure~\ref{fig:views} and details in Table~\ref{table:dental_landmarks}. These landmarks represent key points for subsequent tasks, such as treatment planning. Specifically, for each tooth, the five landmarks identify important anatomical structures that help guide orthodontic measurements and interventions. This meticulous labeling ensures that every model in the dataset captures essential dental characteristics, ranging from normal anatomical variations to more complex anomalies, thereby broadening the scope and applicability of the dataset for diverse research and clinical needs.

\begin{table}[t]
\centering
\caption{Alphanumeric dentition layout.}
\resizebox{\columnwidth}{!}{
\begin{tabular}{|c|c|c|c|c|c|c|c||c|c|c|c|c|c|c|c|}
\hline
\multicolumn{8}{|c||}{Upper Right}  & \multicolumn{8}{c|}{Upper Left} \\ \hline
UR8 & UR7 & UR6 & UR5 & UR4 & UR3 & UR2 & UR1 & LR1 & LR2 & LR3 & LR4 & LR5 & LR6 & LR7 & LR8 \\ \hline \hline
LR8 & LR7 & LR6 & LR5 & LR4 & LR3 & LR2 & LR1 & LL1 & LL2 & LL3 & LL4 & LL5 & LL6 & LL7 & LL8 \\ \hline
\multicolumn{8}{|c||}{Lower Right} & \multicolumn{8}{c|}{Lower Left} \\ \hline
\end{tabular}
}
\label{table:alpha_dentition}
\end{table}

\subsection{Dentition layout and taxonomy} \label{sec:taxonomy}

In this paper, we adopt the alphanumeric notation (see Table \ref{table:alpha_dentition}) to indicate  each tooth. Each tooth is identified by a combination of letters and numbers indicating its position and type. For instance, the upper right first molar is denoted as UR6, while the lower left central incisor is denoted as LL1. This standardized notation facilitates clear communication and precise reference to specific teeth throughout the paper.

The $1,214$ 3D models are divided into different types of dentition to further analyze the performance of the model at different levels of difficulty, something we believe is of utmost importance and missing from previous work. We introduced a taxonomy in which each dentition can be categorized following a two-digit nomenclature that indicates two features. The first digit can be considered a binary variable that indicates if the dentition contains any of the third molars, that is, UR8 and UL8 for the upper dentition and LR8 and LL8 for the lower dentition. The second digit indicates the number of missing teeth (not counting UR8, UL8, LR8 or LL8) in the dentition. Table \ref{table:taxonomy} explains the dentition taxonomy and shows the count in the data set by dentition type. Dental models are centered on the origin and oriented parallel to the $z$ axis. An example of each type of dentition is shown in Figure \ref{fig:examples}. 

\begin{table}[t]
\centering
\caption{Description and number of samples of IOSLandmarks-1k's dentition types.}
\begin{tabular}{llr}
 \hline
 Dent type & Description & \# of 3D dental models \\
 \hline
 00 & No third molar, no missing teeth & 668 \\
 
 01 & No third molar, one missing tooth & 85\\
 
 02 & No third molar, two missing teeth & 106 \\
 
 03 & No third molar, three missing teeth & 14 \\
 
 04 & No third molar, four or more missing teeth & 10 \\
 \hline
 10 & One or two third molars, no missing teeth & 211 \\
 
 11 & One or two third molars, one missing tooth & 44 \\
 
 12 & One or two third molars, two missing teeth & 59 \\
 
 13 & One or two third molars, three missing teeth & 9 \\
 
 14 & One or two third molars, four or more missing teeth & 8 \\
 \hline
\end{tabular}
\label{table:taxonomy}
\end{table}

\begin{figure*}[t]
    \centering
    \includegraphics[width=0.85\textwidth]{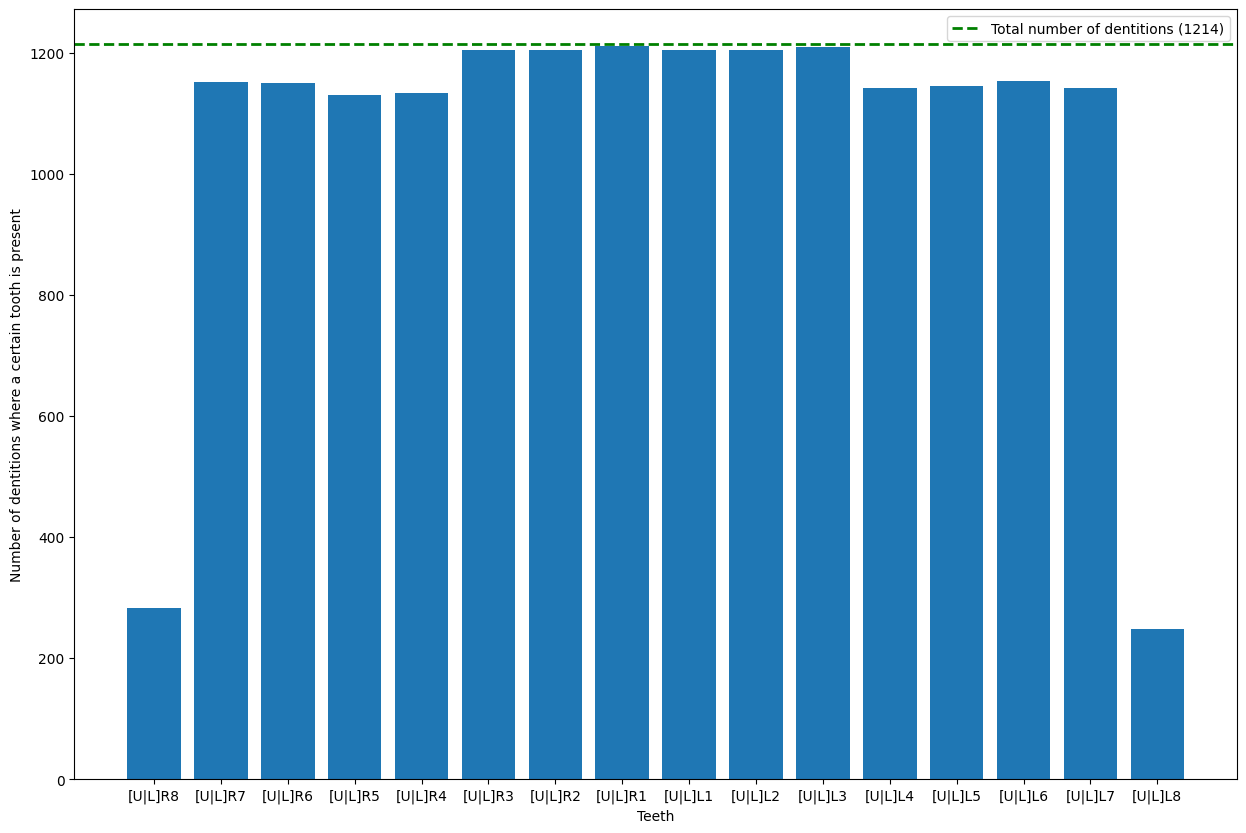}
    \caption{Count of each tooth in the dataset. The green dashed line represents the total number of dental models, i.e., the maximum number of possible appearances of each tooth.}\
    \label{fig:teeth_distribution}
\end{figure*}

\subsection{Teeth distribution}
An important aspect of our dataset is the real-world distribution of teeth across all dental models, which reflects the natural variability observed in clinical practice. Specifically, as shown in Figure \ref{fig:teeth_distribution}, our dataset includes an imbalanced distribution of teeth: less common teeth, such as the third molars, are underrepresented, while more common teeth, such as the first incisors, are almost always present. This imbalance introduces additional challenges for learning methods, as correctly classifying underrepresented teeth becomes more difficult compared to those that appear more frequently.

\begin{figure*}[t]
    \centering
    \includegraphics[width=\textwidth]{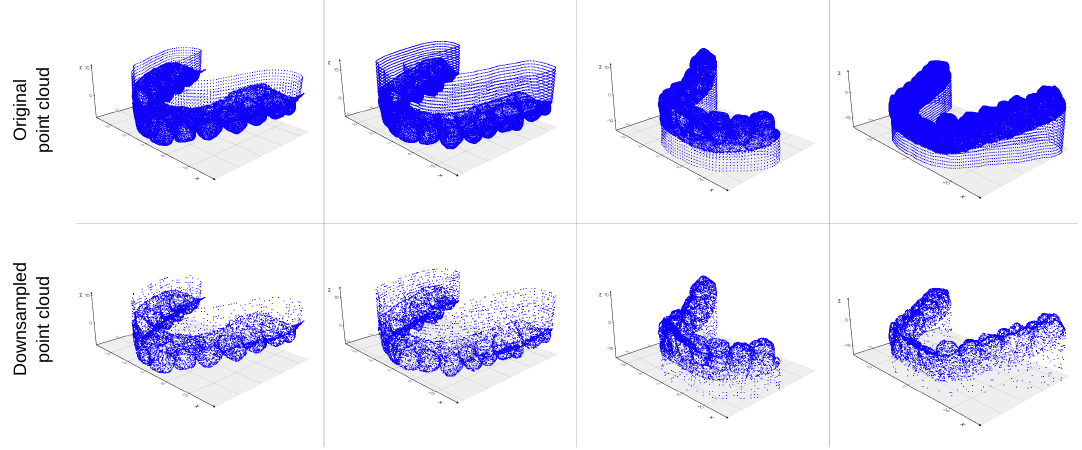}
    \caption{Original (top row) and downsampled (bottom row) point clouds.}\
    \label{fig:downsampling}
\end{figure*}

\begin{figure}[t]
\begin{subfigure}{.48\textwidth}
  \centering
  \includegraphics[width=.8\linewidth]{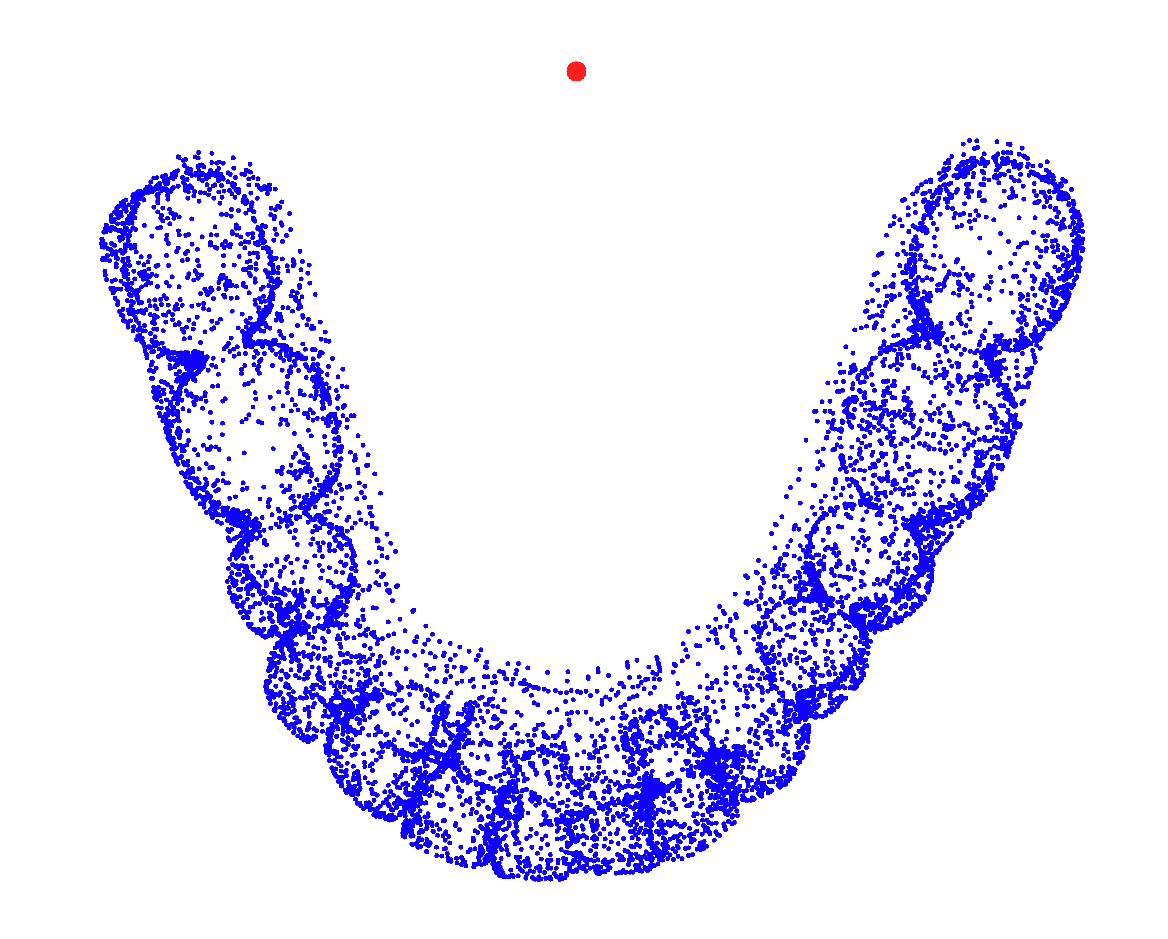}
  \caption{Occlusal view.}
  \label{fig:occlusal_null}
\end{subfigure}%
\begin{subfigure}{.48\textwidth}
  \centering
  \includegraphics[width=.8\linewidth]{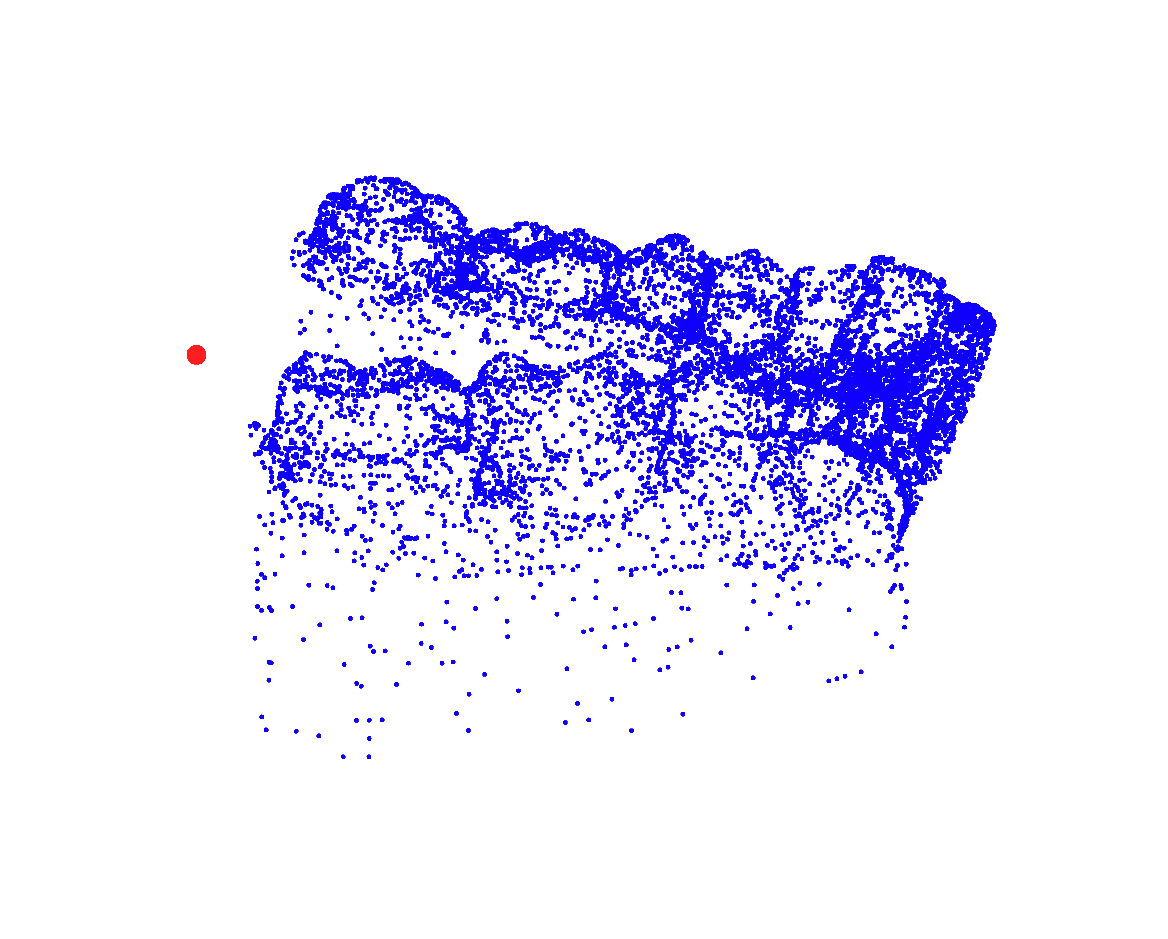}
  \caption{Lateral view.}
  \label{fig:lateral_null}
\end{subfigure}
\begin{subfigure}{.48\textwidth}
  \centering
  \includegraphics[width=.8\linewidth]{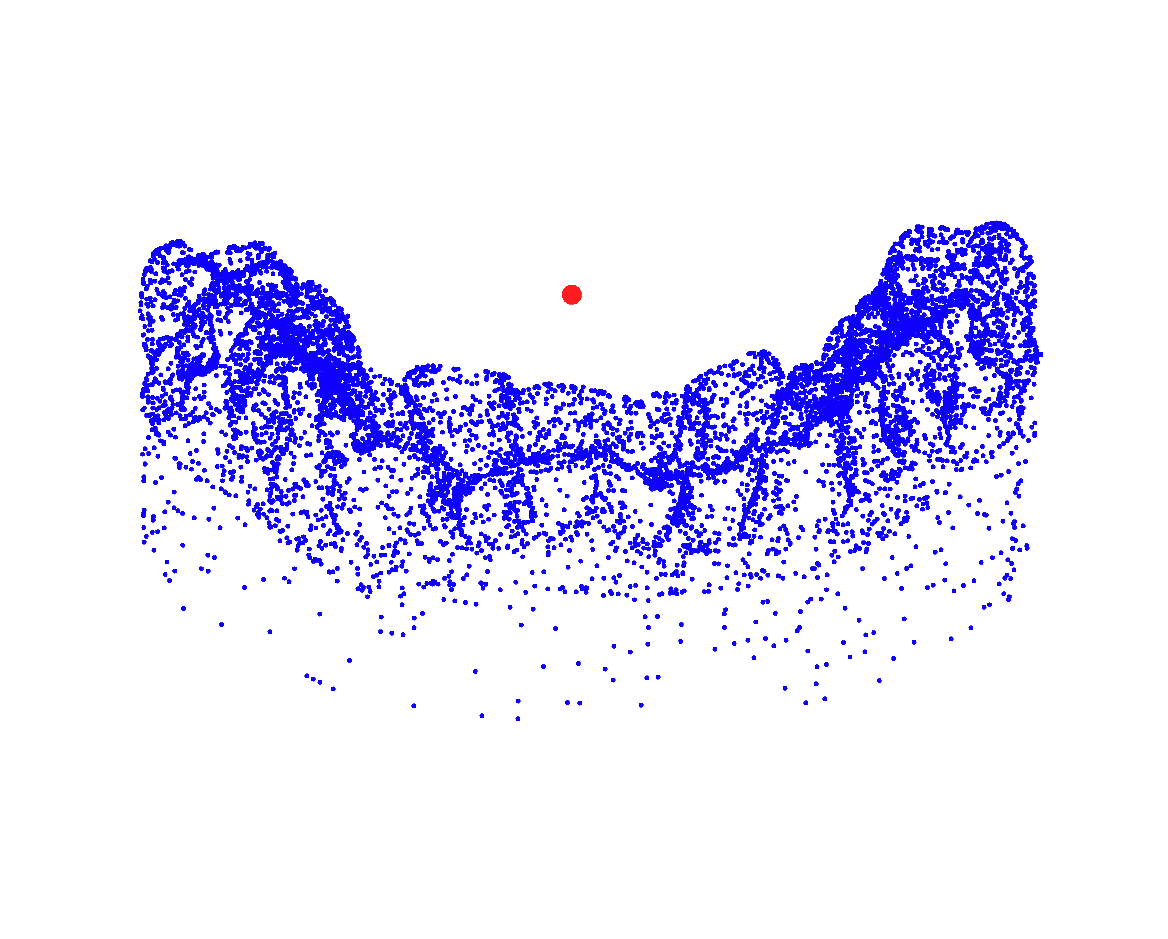}
  \caption{Facial view.}
  \label{fig:facial_null}
\end{subfigure}
\begin{subfigure}{.48\textwidth}
  \centering
  \includegraphics[width=.8\linewidth]{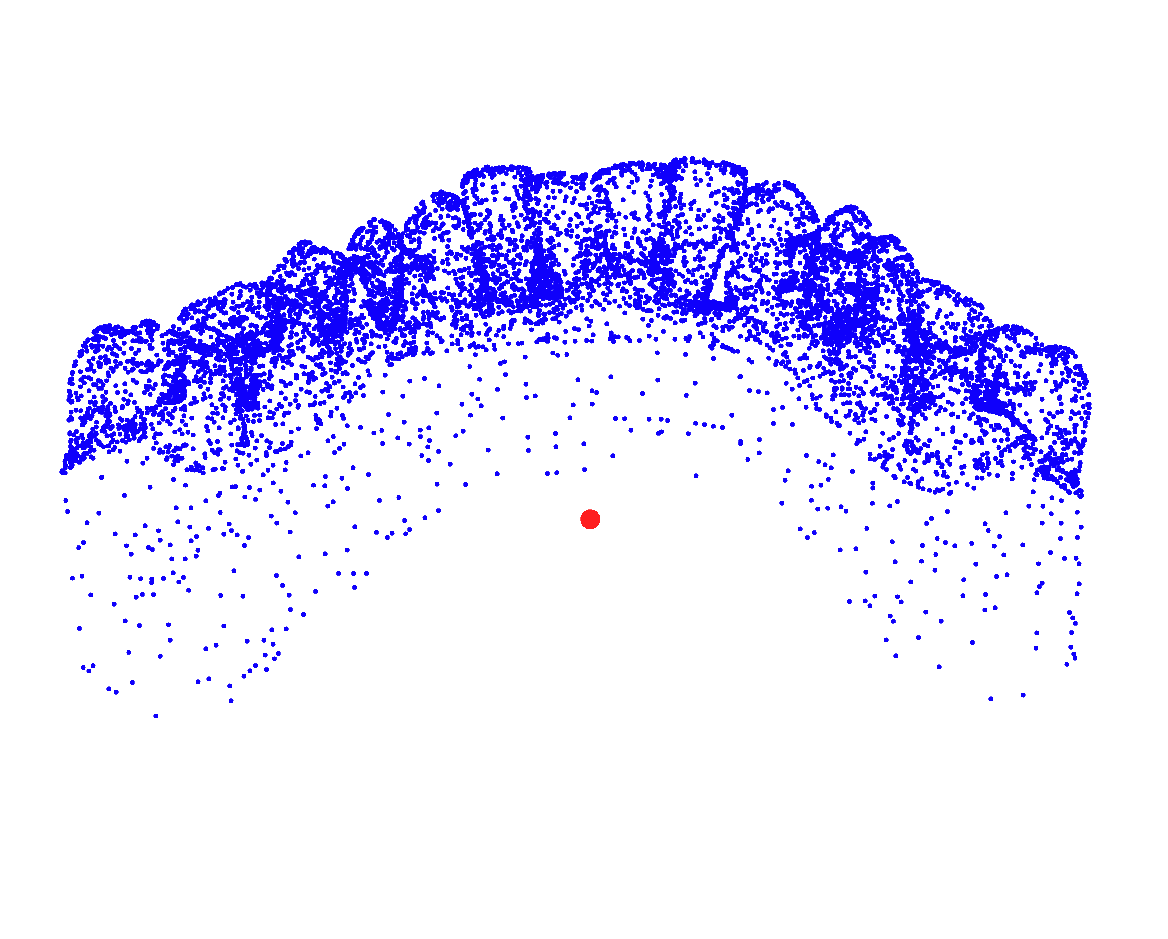}
  \caption{Lingual view.}
  \label{fig:lingual_null}
\end{subfigure}
\caption{4 views of an input sample: A point cloud with 10000 points randomly subsampled from a 3D dental model (blue) plus the added "null" point (red).}
\label{fig:views_null}
\end{figure}

\subsection{Data preprocessing}

In order to feed the data to the point cloud learning methods, our 3D dental models, composed of $67,000$ points and $130,000$ faces on average, are downsampled to $10,000$ vertices, as it allows enough resolution for the point cloud models to learn the desired mapping while reducing memory and computation requirements. Some original and downsampled point clouds are shown in Figure \ref{fig:downsampling}.

Many dental models in the dataset have missing teeth, and since we use a heatmap regression approach for landmark detection, it is necessary to account for landmarks associated with these missing teeth. Allocating these landmarks to any point within the original downsampled point cloud would be illogical, as the corresponding teeth are absent. To address this issue, we introduce an additional point located outside the downsampled point cloud, called the "null point" (see Figure \ref{fig:views_null}). This point serves as a placeholder for the landmarks of missing teeth, ensuring that the model can handle incomplete dental models effectively. The location of the null point is determined in a data-dependent manner and is computed as follows: Let $ m_b $ represent the largest extent of the bounding box enclosing the point cloud across the three axes, and let $c$ denote the centroid of the point cloud. The null point $n$ is then placed at $n = c + \frac{m_b}{2} \cdot (0, 1, 0)$, ensuring it is positioned outside the point cloud along the positive $y$-axis. By adding this point, each point cloud will contain $10001$ vertices in total.

\subsection{Ground-truth generation}

For our model to be able to detect landmarks, we define a Gaussian distance field for each landmark over the point cloud, where higher values are associated with closeness to the corresponding landmark. Namely, for each landmark $l_k$ and for each point in the input point cloud $x_i$, we compute its heatmap value with respect to landmark $k$ as:

\begin{equation}
    h_{ki} = \exp\left(-\frac{d(x_i, l_k)^2}{2\sigma^2}\right)
\end{equation}

where $d$ is the Euclidean distance function, and $\sigma$ is a hyperparameter representing the standard deviation of the Gaussian distance field, which we set to $2 \text{mm}$.

Furthermore, to enable the classification of tooth presence or absence in CHaR versions, we define a \textit{binary label} $y_t$ for each tooth $t$, where $y_t = 1$ indicates the presence of tooth $t$ and $y_t = 0$ indicates its absence. These binary labels are generated based on the dental anatomy provided for each input sample.


\section{Methodology}  \label{sec:methods}

In this section, we first formally define our problem in Section \ref{sec:statement}. Then, our proposed framework is introduced in Section \ref{sec:network}.

\subsection{Problem statement} \label{sec:statement}

Given an input point cloud \( P = \{x_i \in \mathbb{R}^3 \mid i=1, ..., N\} \) (where \( N = 10001 \), including the "null" point), our task is to predict the positions \( L = \{l_k \in \mathbb{R}^3 \mid k = 1, ..., 80\} \) of 80 dental landmarks. Our method employs a heatmap-based approach to encode the likelihood of each point \( x_i \) being a particular landmark. For each landmark \( k \), the heatmap \( \hat{H}_k = \{\hat{h}_{ki} \mid i = 1, ..., N\} \) represents these likelihoods. Then, the predicted landmark positions are defined as:

\begin{equation}
    \hat{L} = \{\hat{l}_k = \underset{x_i \in P}{\operatorname{argmax}}(\hat{H}_k) \mid k=1, ..., 80 \}.
\end{equation}

To train the model, we employ two loss functions: one for the landmark localization task and another for the  tooth presence/absence classification task.

\subsubsection*{Heatmap regression loss}

For the heatmap regression task, we minimize the mean squared error (MSE) loss between the ground truth heatmaps \( H_k = \{h_{ki} \mid i = 1, ..., N\} \) and the predicted heatmaps \( \hat{H}_k \). Specifically, for each landmark \( k \), the MSE loss is computed as:

\begin{equation} \label{eq:mse_loss}
    \mathcal{L}_{\text{MSE}} = \frac{1}{K} \sum_{k=1}^{K} \frac{1}{N} \sum_{i=1}^{N} \left( h_{ki} - \hat{h}_{ki} \right)^2,
\end{equation}

where \( K = 80 \) is the total number of landmarks, and \( N = 10001 \) is the number of points in the point cloud. This loss encourages the model to produce heatmaps that accurately represent the spatial distribution of each landmark, ensuring the predicted landmarks \( \hat{L} \) derived from the maxima of the heatmaps closely match the ground truth landmarks \( L \).

\subsubsection*{Teeth presence classification loss}

For the presence/absence classification task, we define a binary label \( y_t \in \{0, 1\} \) for each tooth \( t \), where \( y_t = 1 \) indicates the presence of the tooth and \( y_t = 0 \) indicates its absence. The model predicts probabilities \( \hat{y}_t \in [0, 1] \) for each tooth \( t \), and we optimize these predictions using the binary cross-entropy (BCE) loss:

\begin{equation} \label{eq:bce_loss}
    \mathcal{L}_{\text{BCE}} = -\frac{1}{T} \sum_{t=1}^{T} \left[ y_t \log(\hat{y}_t) + (1 - y_t) \log(1 - \hat{y}_t) \right],
\end{equation}

where \( T \) is the total number of teeth under consideration. This loss ensures that the model accurately predicts the presence or absence of teeth.

\subsubsection*{Combined loss function}

To jointly train the model for landmark localization and tooth presence/absence classification, we define a combined loss function with separate weights for each task:

\begin{equation} \label{eq:combined_loss}
    \mathcal{L} = \lambda_{\text{reg}} \mathcal{L}_{\text{MSE}} + \lambda_{\text{cls}} \mathcal{L}_{\text{BCE}},
\end{equation}

where \( \lambda_{\text{reg}} \) and \( \lambda_{\text{cls}} \) are hyperparameters that control the relative importance of the localization task (\( \mathcal{L}_{\text{MSE}} \)) and the classification task (\( \mathcal{L}_{\text{BCE}} \)), respectively.

By minimizing this combined loss, the model learns to simultaneously predict accurate landmark positions and classify the presence or absence of teeth, ensuring robust performance across both tasks.

\subsection{CHaRM} \label{sec:network}
\begin{figure*}[t]
    \centering
    \includegraphics[width=\textwidth]{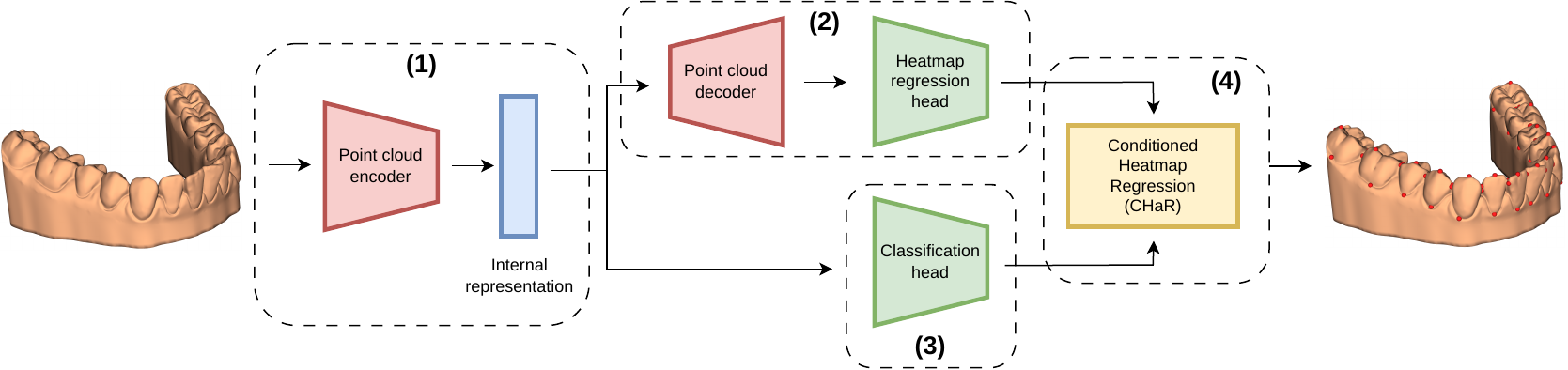}
    \caption{Pipeline of CHaRM. Given an input point cloud of a 3D dental model, (1) the point cloud encoder extracts internal features, which are then simultaneously processed by two parallel branches: (2) a landmark regression branch that generates initial heatmaps for each landmark, and (3) a teeth presence classification branch that predicts the existence of each tooth. Finally, (4) the Conditioned Heatmap Regression (CHaR) module uses the presence predictions to refine the initial landmark heatmaps.  Point cloud encoder and decoder (shown in red) can be any point cloud learning architecture, such as PointMLP or PointTransformer.}
    \label{fig:flowchart}
\end{figure*}

In this paper, we propose CHaRM (Conditioned Heatmap Regression Methodology), an end-to-end DL-based landmark detection methodology for 3D dental models, which does not rely on previous segmentation of teeth. It is able to handle missing teeth and fasten inference, making it suitable for clinical applications. Our proposed methodology is shown in Figure \ref{fig:flowchart}, where (1) the point cloud encoder extracts the internal representation, (2) the point cloud decoder plus the heatmap regression head performs the initial heatmap-based regression of landmarks, (3) the classification head predicts the presence or absence of every tooth, and (4) our novel conditioned heatmap regression module adjusts every heatmap's scores based on the presence classification of its associated teeth.

\begin{figure*}[t]
    \centering
    \includegraphics[width=\textwidth]{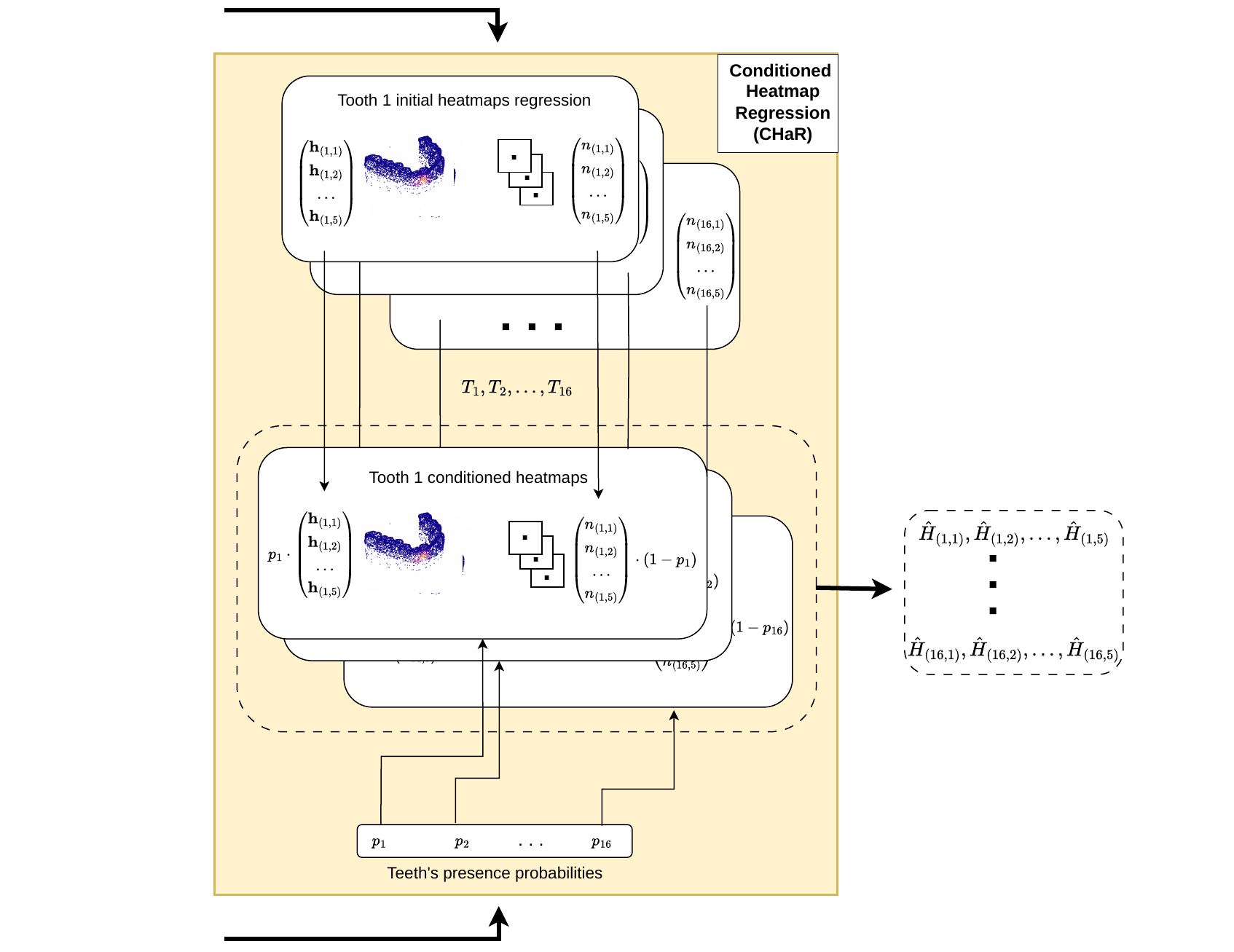}
    \caption{Overview of the Conditioned Heatmap Regression (CHaR) module. Initial heatmaps are adjusted by weighting the original point cloud likelihoods with the predicted tooth presence probabilities $p_t$ and the null point likelihoods with the absence probabilities $(1-p_t)$. }
    \label{fig:conditioned}
\end{figure*}
 
\subsubsection*{Encoder} 
The point cloud encoder is in charge of extracting a reduced internal representation of the whole input point cloud. Point cloud learning methods usually perform this dimensionality reduction and feature selection hierarchically through set abstraction operations \cite{qi2017pointnet++, qian2022pointnext}. These learned high-level features will subsequently be fed to the decoder plus heatmap regression head and the teeth presence classification.

\subsubsection*{Decoder + heatmap regression}
The decoder reverses the encoding process by progressively interpolating the abstracted features back to the original number of points in the input point cloud. This step ensures that the high-level features extracted by the encoder are effectively propagated back to the spatial resolution of the input. Subsequently, the heatmap regressor computes the initial likelihood that each point in the point cloud is associated with a specific landmark.

This is achieved by generating a set of heatmaps, where each heatmap corresponds to a landmark and encodes the likelihood distribution over all points in the point cloud. Formally, for each landmark \( k \) we obtain the initial heatmap \( \hat{H}_k' = \{\hat{h}_{ki}' \mid i = 1, ..., N\} \), which will be used to compute the final heatmap \( \hat{H}_k
\) as  explained below.

\subsubsection*{Teeth presence classification}

The classification head of our network employs a multi-layer sequential architecture for robust feature extraction and class prediction. It takes the high-level features extracted by the encoder as input and processes them through three fully connected layers, each interleaved with batch normalization, ReLU activation, and dropout for regularization. The first two layers project the input into a 256-dimensional feature space, stabilizing training with batch normalization and introducing non-linearity through ReLU activations. Dropout is applied after each activation to mitigate overfitting. The final layer maps the features into an output space of 16 classes, corresponding to the number of teeth in a dental arch, producing the desired logits. A sigmoid function is then applied to these logits, generating a probability vector $\mathbf{p} = [p_1, p_2, ..., p_{16}]$, where each $p_i$ represents the likelihood of the presence of a corresponding tooth. These probabilities are subsequently used to condition the final heatmap regression.

\subsubsection*{Conditioned Heatmap Regression (CHaR)}

In general, when a tooth is missing, the vanilla landmark regression network is not explicitly designed to map all landmarks associated with that tooth to the null point. To address this issue, the conditioned heatmap regression (CHaR) module adjusts the initial landmark regression to take into account the predicted presence probability of each tooth. For clarity, each landmark $ k $ is denoted using a tuple of two digits $(t,g)$, where the first digit \( t = 1, 2, ..., 16 \) indicates the tooth associated with the landmark, and the second digit $ g = 1, 2, ..., 5 $ indicates the specific landmark within the group of 5 landmarks associated with tooth $ t $. For example, landmark $ 16 $ (the 1st landmark of the 4th tooth) is denoted as $ (4, 1)$ in this notation.

The intuition behind the CHaR module is to condition the likelihood values of each landmark $ (t, g) $ on the predicted presence probability $ p_t $ of the associated tooth. Specifically, for each landmark $ (t, g) $, the likelihood of every point in the original point cloud (excluding the null point) is weighted by $ p_t $, while the null point is weighted by the absence probability $ (1 - p_t) $. Formally, assuming that the null point is the last point in the input point cloud, the final heatmap values for landmark $(t, g)$ are computed as:

\begin{equation}
\hat{H}_{t,g} = \{h_{(t,g)i}' \cdot p_t \mid i = 1, ..., N-1\} \cup \{h_{(t,g)N}' \cdot (1-p_t)\}
\end{equation}

In this formulation, the weights \( p_t \) and \( (1 - p_t) \) act as a switch: when \( p_t \) is high (tooth \( t \) is present), the likelihoods of points in the original point cloud dominate, encouraging the landmarks associated with tooth \( t \) to remain in the point cloud. In contrast, when \( p_t \) is low (tooth \( t \) is absent), the probability of a null point dominates, shifting the associated landmarks to the null point. Figure \ref{fig:conditioned} provides a schematic representation of our CHaR module. For clarity, we denote the set of likelihood values for the original points (excluding the null point) as $\mathbf{h}_{(t,g)} = \{h_{(t,g)i}' \mid i = 1, ..., N-1\}$, and the likelihood value for the null point as $n_{(t,g)} = h_{(t,g)N}'$.

This conditioning ensures that the network dynamically adjusts the landmark predictions based on the presence of the tooth, improving the robustness in scenarios where some teeth are missing.

\section{Experiments} \label{sec:experiments}


In this section, we show the effectiveness of CHaRM by coupling the CHaR module with several point cloud learning models spanning different families of algorithms and compare the best CHaR-based network (CHaRNet) with SOTA landmark detection models. The dataset splitting procedure is explained in Section \ref{sec:data_split}. The experimental setup of each compared model is shown in \ref{sec:exp_setup}. A description of the evaluation protocol used is introduced in Section \ref{sec:eval_protocol}. A comparison between baseline point cloud models and their corresponding CHaR variants is presented in Section \ref{sec:results_char}. Finally, the comparison between CHaRNet and SOTA methods is introduced in Section \ref{sec:CHaR_SOTA}.

\subsection{Dataset splitting} \label{sec:data_split}

We follow the typical three-way holdout setup for model comparison and performance estimation \cite{Raschka_2020}. Specifically, the 1,214 dental models are divided into three subsets: training (70\%), validation (15\%), and testing (15\%). To prevent patient-level correlation between subsets, we first group dental models by patient, ensuring that all models from a single patient belong exclusively to one of these subsets. Furthermore, to guaranty balanced representation of different anatomical complexities across these subsets, the splitting is conducted in a stratified manner according to the dental type taxonomy defined in Section \ref{sec:taxonomy}. This stratification ensures that the distribution of dentition types remains consistent across training, validation, and testing subsets, enabling reliable evaluation of model performance on diverse dental anatomies.

\subsection{Experimental setup} \label{sec:exp_setup}


The experimental setup for each compared method in this work are as follows: 

\begin{itemize}
    \item CHaR-based networks: The different CHaR-based versions were trained by minimizing the MSE loss (Formula \ref{eq:mse_loss}) and the combined loss (Formula \ref{eq:combined_loss}), respectively, using the Adam optimizer \cite{kingma2014adam} for a total of 100 epochs and a batch size of 16. The initial learning rate is set to 0.005 and is progressively reduced using the cosine learning rate decay scheduler. To prevent overfitting, weight decay and dropout rate were set to 0.003 and 0.5, respectively, serving as regularization terms.  With respect to hyper-parameters controlling the relative importance in the final loss value in the CHaR versions, $\lambda_{\text{reg}}$ and $\lambda_{\text{cls}}$ are set to 0.001 and 1 respectively, given the natural bigger quantity of the landmark localization loss. 
    \item TSMDL: Teeth segmentation network, iMeshSegNet, and landmark regression network, PointNet-Reg, are trained in IOSLandmarks-1k using default parameters. 
    \item ALIIOS: iMeshSegNet is used as teeth segmentation method since ALIIOS' original teeth segmentation network, DentalModelSeg, is not available to be trained in IOSLandmarks-1k. We also change the default camera settings to accommodate the cameras' positioning to our landmarks of interest described in Table \ref{table:dental_landmarks}.
\end{itemize}

All methods were trained in the same server with Intel Xeon E5-2698 as CPU and NVIDIA Tesla V100 32Gb as GPU.

\subsection{Evaluation protocol} \label{sec:eval_protocol}

To rigorously evaluate landmark detection methods on dental models, especially when certain teeth (and thus their landmarks) might be absent, we evaluate the model in two clinically meaningful tasks: landmarks/tooth presence detection and landmark localization. 

\subsubsection*{Landmarks/Tooth presence detection}

We first evaluate the model's ability to correctly determine the presence or absence of each tooth, which is equivalent to determining the presence or absence of each landmark since a set of five landmarks is associated with a tooth. Then, for every landmark prediction, four distinct outcomes are possible:

\begin{itemize}
    \item True Positive (TP): Landmark exists in the mesh and is correctly predicted in the mesh.
    \item False Positive (FP): Landmark does not exist but is erroneously predicted in the mesh.
    \item False Negative (FN): Landmark exists in the mesh but is erroneously predicted at the null point.
    \item True Negative (TN): Landmark does not exist and is correctly predicted at the null point.
\end{itemize}

We compute the F1-score to evaluate landmarks/tooth classification performance: 

\begin{equation} \label{eq:f1_l}
        F1 = \frac{2*TP}{2*TP+FP+FN}
\end{equation}

\subsubsection*{Landmark localization}
For landmarks correctly identified as present (true positives), we measure how accurately the model predicts their 3D coordinates. Let $D_{i} = \|\hat{\mathbf{l}}_i - \mathbf{l}_i\|_2$ be the distance between the predicted and ground-truth landmark positions. To assess this accuracy, we employ two metrics:

\begin{itemize}
    \item Mean Euclidean Distance Error (MEDE): This metric calculates the average Euclidean distance between predicted and ground-truth landmark positions, computed exclusively over the true positives: 
    \begin{equation}
    \text{MEDE} = \frac{1}{|TP|}\sum_{i \in TP} D_{i}
    \end{equation}

    \item Mean Success Ratio (MSR): To measure localization robustness, we compute the MSR at a clinically relevant threshold of 1 mm. MSR represents the percentage of true-positive landmark predictions falling within 1 mm of their ground-truth locations:
    \begin{equation}
       \text{MSR} =\frac{1}{|TP|}\sum_{i \in TP} \delta_{i} \times 100\%, \quad
        \delta_{i} = \begin{cases} 
              1 & \text{if } D_{i} \leq r, \\
              0 & \text{if } D_{i} > r
           \end{cases}
    \end{equation}
\end{itemize}



\subsubsection*{Motivation for the evaluation protocol}
The motivation for this protocol arises from the clinical reality that landmark-presence detection and landmark localization are fundamentally different tasks. The presence detection reflects a classification problem (tooth existence), while localization assesses spatial accuracy once the existence is confirmed. Evaluating Euclidean distances involving the artificial "null point" (used to handle absent landmarks) would introduce arbitrary distance values with no clinical significance, thus introducing noise in the true localization performance evaluation. Our proposed protocol resolves this by evaluating localization accuracy strictly on landmarks correctly identified as present, ensuring both metrics remain clinically meaningful and transparent.

\subsection{Base versus CHaR-based point cloud models} \label{sec:results_char}

\begin{table}[t]
\centering
\caption{Comparison of each neural network base model with its corresponding Conditioned Heatmap Regression (CHaR) version in terms of F1-score for the task of teeth classification. The results are expressed across each dentition type and aggregated using the macro and micro averaged F1-score. \textbf{Bold} indicates the best F1-score per model.}
\resizebox{\columnwidth}{!}{%
\begin{tabular}{l r r r r r r r r r r}
\hline
 & \multicolumn{10}{c}{Models} \\
\hline
 & \multicolumn{2}{c}{PointNet++} & \multicolumn{2}{c}{DGCNN} & \multicolumn{2}{c}{PointTransformer} & \multicolumn{2}{c}{PointNeXt} & \multicolumn{2}{c}{PointMLP} \\ 
\hline
Dent type &  Base & CHaR  & Base & CHaR & Base & CHaR & Base & CHaR & Base & CHaR\\
\hline
00 &  \textbf{0.95} & 0.87 & 0.87 &  \textbf{0.99} & 0.93 & \textbf{0.99} & 0.93 & \textbf{0.99} & 0.93 & \textbf{0.99} \\
01 &  0.84 & \textbf{0.98} & 0.84 &  \textbf{0.97} & 0.85 & \textbf{0.92} & 0.85 & \textbf{0.86} & 0.91 & \textbf{0.98} \\
02 & 0.93 & \textbf{0.96}   & 0.79 &   \textbf{0.95} & 0.80 & \textbf{0.97} & 0.87 & \textbf{0.95} & 0.81 & \textbf{0.95} \\
03 &  0.89 & \textbf{0.93} & 0.87 &  \textbf{0.93} & \textbf{0.91} & 0.89 & 0.91 & 0.91 & \textbf{0.89} & 0.85 \\
04 &  0.87 & \textbf{0.93} & 0.68 &  \textbf{0.87} & 0.87 & \textbf{0.93} & 0.81 & 0.81 & 0.81 & \textbf{0.93} \\
10 &  0.99 & 0.99 & 0.99 &  0.99 & 0.97 & \textbf{0.99} & 0.97 & \textbf{0.99} & 0.99 & 0.99 \\
11 & 0.94 & 0.95 &  0.93 & \textbf{0.95} & 0.97 & 0.97 & 0.94 & \textbf{0.96} & 0.95 & \textbf{0.98} \\
12 & 0.91 & \textbf{0.94} & 0.91 & \textbf{0.91} & 0.90 & \textbf{0.93} & 0.91 & \textbf{0.93} & \textbf{0.93} & 0.92 \\
13 & 0.81 & \textbf{0.87} & 0.75 &  0.75 & 0.75 & \textbf{0.93} & 0.81 & \textbf{0.87} & 0.87 & \textbf{0.90} \\
14 & 0.56 &  \textbf{0.68} &  0.56 & \textbf{0.62} & 0.62 & \textbf{0.81} & 0.62 & \textbf{0.68} & 0.62 & \textbf{0.68} 
\\

Macro-avg & 0.87 & \textbf{0.91} & 0.82 &  \textbf{0.89} & 0.86 & \textbf{0.93} & 0.86 & \textbf{0.90} & 0.87 & \textbf{0.92} \\
Micro-avg & 0.97 & \textbf{0.98} & 0.96 &  \textbf{0.98} & 0.95 & \textbf{0.98} & 0.98 & \textbf{0.98} & 0.98 & \textbf{0.99} \\
\hline
\end{tabular}
}
\label{table:f1}
\end{table}

We compare five base point cloud models (PointNet++, DGCNN, PointTransformer, PointNeXt, and PointMLP) against their CHaR-based versions along two complementary sources of error: (i) tooth/landmarks presence classification in Table \ref{table:f1}, and (ii) landmark localization accuracy and robustness, only for the TP landmarks, in Table \ref{table:mede} for MEDE and Table \ref{table:msr} for MSR. Because only the CHaR-based versions have an explicit tooth presence output, for the base models we infer tooth presence heuristically: a tooth is considered present if all five of its associated landmarks are predicted inside the mesh; if any of them is sent to the null point, the tooth is treated as absent. This allows a fair, consistent definition of true positives for the localization metrics.

Table \ref{table:f1} shows that CHaR augmentation consistently improves the F1-score for tooth presence across architectures, with the largest gains in anatomically irregular cases (missing teeth / third molars). Better presence classification reduces the propagation of existence errors into localization, laying the foundation for more reliable downstream predictions.

When measuring localization performance, Tables \ref{table:mede} and \ref{table:msr} show that CHaR-based models generally improve spatial precision (lower MEDE) and increase the fraction of clinically acceptable predictions (higher MSR at 1 mm) across all point cloud models. The CHaR-based PointMLP model, which we call CHaRNet, exhibits the strongest localization profile: it achieves the lowest MEDE and highest MSR in the majority of dentition types, indicating both tighter errors and a larger fraction of landmarks falling within the clinically relevant 1 mm radius. For example, CHaRNet’s MSR improvement over its base counterpart is consistent across standard and challenging dentitions, reflecting robustness to anatomical variability. The reduction in MEDE demonstrates that, beyond simply detecting which landmarks to localize, CHaRNet refines their spatial estimates, producing more accurate coordinate predictions when the tooth is correctly recognized.

The improvements are especially meaningful in intermediate-to-hard cases where partial structure and missing teeth introduce uncertainty. In these settings, the conditioning provided by the CHaR module helps disambiguate the heatmap regression by weighting likelihoods according to tooth presence probabilities, effectively concentrating probability mass on the dentition. This leads to both fewer gross mislocalizations (reflected in MSR gains) and smaller average distance error (lower MEDE). That said, in extremely incomplete dentitions where the presence classification degrades, the conditioning can sometimes mislead localization, which is seen as occasional increases in MEDE for certain subclasses. This underscores that the efficacy of CHaRNet depends on reasonably clear teeth presence geometric signals.

\begin{figure*}[t]
    \centering
    \includegraphics[width=1\textwidth]{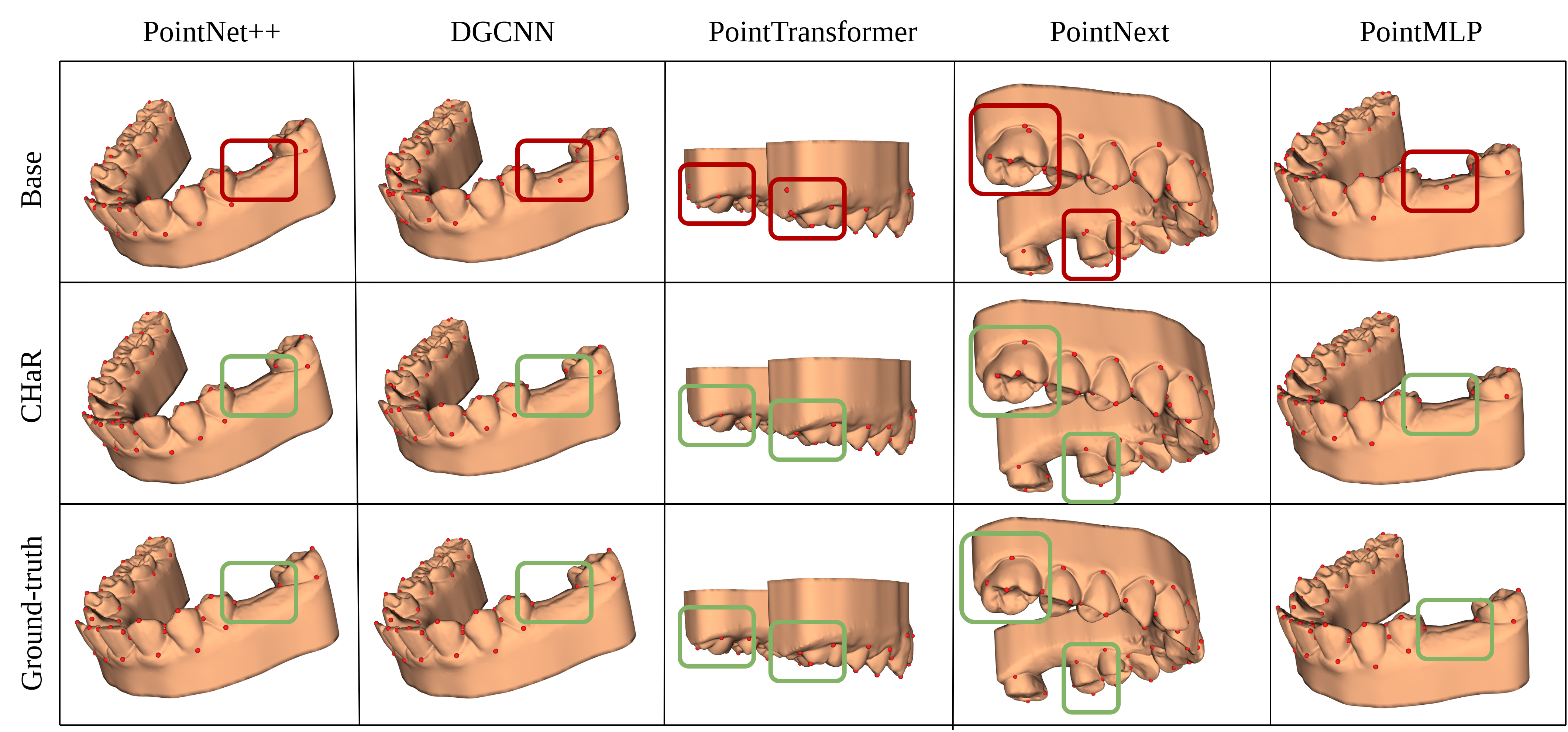}
    \caption{Qualitative comparison between base networks (top row), CHaR-based networks (middle row) and ground truth (bottom row). Each column represents one of the point cloud learning models. The red and green squares represents wrong and correct landmark localizations, respectively.}
    \label{fig:failures}
\end{figure*}

Furthermore, qualitative results between the base network, CHaR-based network, and the ground truth for different point cloud methods and dental models are shown in Figure \ref{fig:failures}. The red and green squares show wrong and correct landmark localization, respectively. Base models struggle to detect/identify missing teeth on incomplete dental geometries by themselves. CHaR-based models tend to do a much better job across these complex anatomies.

In conclusion, CHaRM couples improved landmarks/tooth presence detection with more focused and reliable landmark regression. The addition of the CHaR module yields substantial gains in both precision and robustness, making it better suited for realistic clinical scenarios with variable and incomplete anatomy.


\begin{table}[t]
\centering
\caption{Comparison of each neural network base model with its corresponding Conditioned Heatmap Regression (CHaR) version in terms of mean Euclidean distance error (MEDE) expressed in millimeters (mm). The results are expressed across each dentition type and aggregated using the macro and micro averaged MEDE. \textbf{Bold} indicates the best MEDE per model.}
\resizebox{\columnwidth}{!}{%
\begin{tabular}{l r r r r r r r r r r}
\hline
 & \multicolumn{10}{c}{Models} \\
\hline
 & \multicolumn{2}{c}{PointNet++} & \multicolumn{2}{c}{DGCNN} & \multicolumn{2}{c}{PointTransformer} & \multicolumn{2}{c}{PointNeXt} & \multicolumn{2}{c}{PointMLP} \\ 
\hline
Dent type &  Base & CHaR  & Base & CHaR & Base & CHaR & Base & CHaR & Base & CHaR\\
\hline
00 &  0.90 & 0.90 & 0.83 &  \textbf{0.82} & 0.75 & \textbf{0.65} & 0.85 & \textbf{0.80} & 0.73 & \textbf{0.56} \\
01 &  1.47 & \textbf{1.40} & 1.34 &  \textbf{1.22} & 0.88 & \textbf{0.87} & \textbf{1.10} & 1.15 & 1.10 & \textbf{0.76} \\
02 & \textbf{1.33} & 1.44   & 1.51 &  1.51 & 1.29 & \textbf{0.99} & 1.30 & \textbf{1.25} & 1.18 & \textbf{0.99} \\
03 &  2.68 & \textbf{1.91} & \textbf{2.04} &  2.20 & \textbf{0.89} & 1.15 & \textbf{1.30} & 1.38 & \textbf{1.89} & 1.66 \\
04 &  3.33 & \textbf{3.26} & 3.54 &  \textbf{3.48} & 4.77 & \textbf{2.71} & \textbf{2.68} & 2.76 & 3.85 & \textbf{2.14} \\
10 &  0.98 & \textbf{0.97} & \textbf{0.82} &  0.85 & 0.83 & \textbf{0.73} & 0.91 & \textbf{0.86} & 0.82 & \textbf{0.64} \\
11 & 2.04 & \textbf{1.84} &  1.83 & 1.83 & 1.12 & 1.12 & 1.76 & \textbf{1.53} & 0.97 & \textbf{0.98} \\
12 & \textbf{1.47} & 1.52 &  1.67 & \textbf{1.41} & 0.97 & \textbf{0.84} & 1.28 & \textbf{1.14} & 0.83 & \textbf{0.67} \\
13 & 2.27 & \textbf{1.31} & 1.20 &  \textbf{0.95} & 3.31 & \textbf{0.92} & 1.27 & \textbf{1.18} & \textbf{0.98} & 0.85 \\
14 & 4.54 &  \textbf{3.60} &  5.21 & \textbf{4.61} & \textbf{1.42} & 1.92 & 4.58 & \textbf{3.56} & 2.75 & \textbf{1.91} 
\\

Macro-avg & 2.10 & \textbf{1.82} & 2.00 &  \textbf{1.89} & 1.62 & \textbf{1.19} & 1.70 & \textbf{1.56} & 1.51 & \textbf{1.12} \\
Micro-avg & 1.15 & \textbf{1.08} & 1.02 &  \textbf{1.01} & 0.88 & \textbf{0.76} & 0.99 & \textbf{0.95} & 0.85 & \textbf{0.70} \\
\hline
\end{tabular}
}
\label{table:mede}
\end{table}

\begin{table}[t]
\centering
\caption{Comparison of each neural network base model with its corresponding Conditioned Heatmap Regression (CHaR) version in terms of mean success ratio (MSR) expresed in \% of landmarks located withing a sphere of $1mm$ radius. The results are expressed across each dentition type and aggregated using the macro and micro averaged MSR. \textbf{Bold} indicates the best MSR per model.}
\resizebox{\columnwidth}{!}{%
\begin{tabular}{l r r r r r r r r r r}
\hline
 & \multicolumn{10}{c}{Models} \\
\hline
 & \multicolumn{2}{c}{PointNet++} & \multicolumn{2}{c}{DGCNN} & \multicolumn{2}{c}{PointTransformer} & \multicolumn{2}{c}{PointNeXt} & \multicolumn{2}{c}{PointMLP} \\ 
\hline
Dent type &  Base & CHaR  & Base & CHaR & Base & CHaR & Base & CHaR & Base & CHaR\\
\hline
00 &  63.9 & \textbf{64.4} & 74.0 & \textbf{74.2} & 74.8 & \textbf{81.1} & 66.1 & \textbf{73.7} & 75.2 & \textbf{85.2} \\
01 &  \textbf{47.3} & 47.1 & 62.9 & \textbf{62.9} & 70.2 & \textbf{73.8} & 57.0 & \textbf{63.4} & 67.1 & \textbf{77.6} \\
02 & 50.0 & \textbf{52.3} & 61.3 & \textbf{62.8} & 63.7 & \textbf{73.4} & 55.0 & \textbf{65.7} & 64.1 & \textbf{79.5} \\
03 &  25.7 & \textbf{36.2} & \textbf{45.0} & 40.7 & 69.4 & \textbf{69.6} & 45.0 & \textbf{55.6} & 43.0 & \textbf{49.1} \\
04 &  6.6 & \textbf{11.1} & 11.1 &  \textbf{24.4} & 9.3 & \textbf{40.0} & 15.5 & \textbf{31.2} & 20.0 & \textbf{29.5} \\
10 &  58.6 & \textbf{59.4} & \textbf{72.0} &  71.4 & 69.2 & \textbf{75.8} & 61.9 & \textbf{66.7} & 69.9 & \textbf{79.7}  \\
11 & \textbf{41.7} & 40.0 &  52.9 & \textbf{54.5} & 59.2 & \textbf{64.5} & 46.0 & \textbf{53.3} & 62.2 & \textbf{70.9} \\
12 & \textbf{51.7} & 49.7 &  60.9 & \textbf{65.7} & 63.8 & \textbf{72.9} & 50.5 & \textbf{56.3} & 69.5 & \textbf{80.1} \\
13 & 21.0 & \textbf{38.3} & 46.5 &  \textbf{60.0} & 30.1 & \textbf{61.6} & 39.6 & \textbf{48.7} & 55.0 & \textbf{69.0} \\
14 & 20.0 & \textbf{31.1} &  33.3 & \textbf{35.5} & 61.2 & \textbf{66.6} & 18.0 & \textbf{23.7} & 48.8 & \textbf{63.8} 
\\
Macro-avg & 38.7 & \textbf{42.9} & 52.0 &  \textbf{55.2} & 57.1 & \textbf{67.9} & 45.5 & \textbf{53.8} & 58.4 & \textbf{68.5} \\
Micro-avg & 55.5 & \textbf{57.8} & 66.7 &  \textbf{68.9} & 69.2 & \textbf{75.8} & 60.6 & \textbf{68.4} & 70.1 & \textbf{79.8} \\
\hline
\end{tabular}
}
\label{table:msr}
\end{table}

\subsection{CHaRNet versus SOTA} \label{sec:CHaR_SOTA}

We benchmark CHaRNet against the SOTA methods for IOS landmarks detection, TSMDL \cite{wu2022two} and ALIIOS \cite{baquero2022automatic}, using the evaluation method described in Section \ref{sec:eval_protocol}. For segmentation-based methods like TSMDL and ALIIOS, tooth presence classification is derived from the output of the teeth segmentation stage. Note that TSMDL and ALIIOS use the same teeth segmentation method, so results of tooth classification performance are the same. To ensure experimental fairness, all three methods were trained until convergence and evaluated using the same dataset split described in Section \ref{sec:data_split}. The same hardware (CPU: Intel Core i7-13700H; GPU: NVIDIA RTX 4050) and the same batch size (1) were used to compute inference times in Table \ref{tab:execution_times}.

\begin{figure}[t]
     \centering
     \begin{subfigure}[b]{0.85\textwidth}
         \centering
         \includegraphics[width=\textwidth]{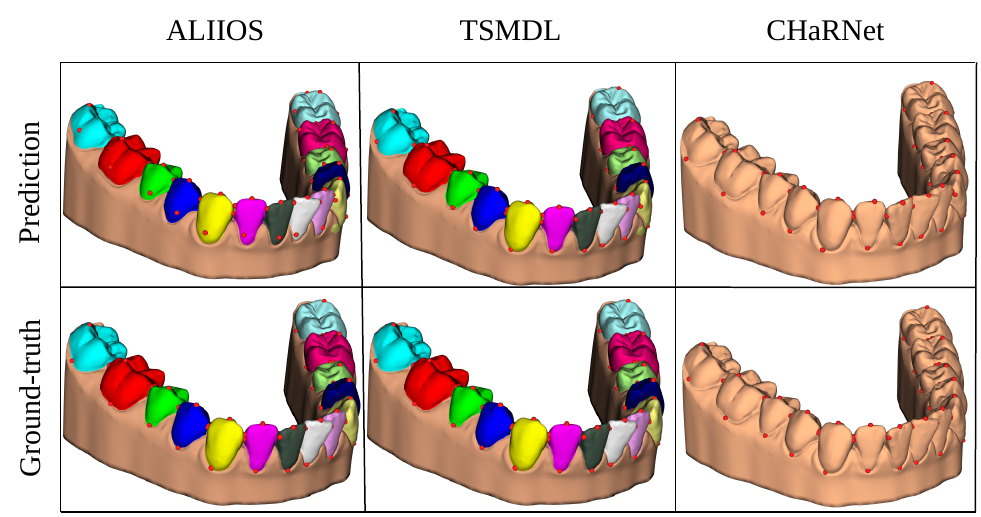}
         \caption{Standard dentition (00)}
         \label{fig:simple_dent}
     \end{subfigure}
     \hfill
     \begin{subfigure}[b]{0.85\textwidth}
         \centering
         \includegraphics[width=\textwidth]{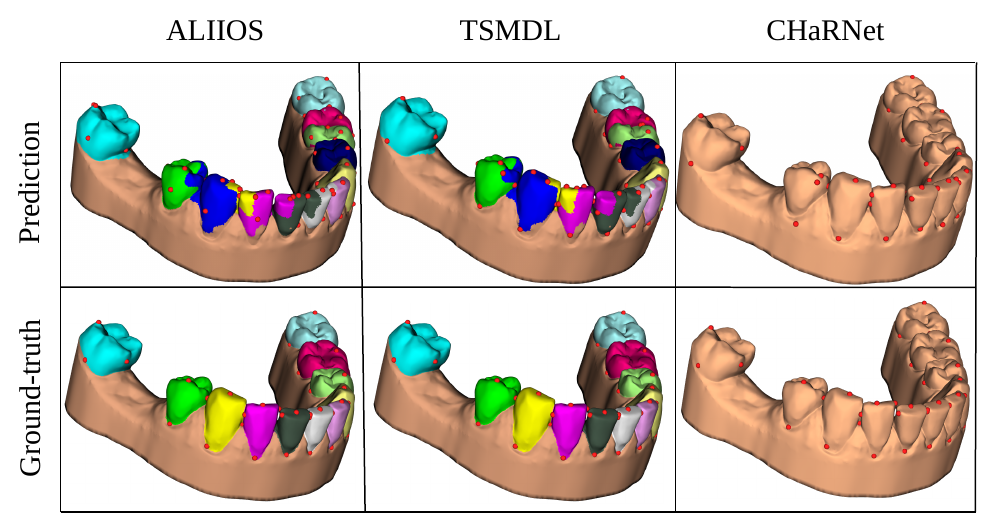}
         \caption{Complex dentition (03)}
         \label{fig:complex_dent}
     \end{subfigure}
        \caption{Qualitative comparison between ALIIOS, TSMDL and CHaRNet on a standard dentition (a) and on a complex dentition (b). Predicition (top row) and ground-truth (bottom row) are shown for each model. Note that segmentation predictions and ground-truth are also provided for ALIIOS and TSMDL.}
        \label{fig:qualitative_SOTA}
\end{figure}

Table \ref{tab:SOTA_f1} shows the overall superior performance in the tooth classification task of CHaRNet versus TSMDL and ALIIOS. CHaRNet achieves the best F1-score across all dentition types, macro and micro averages, which further improves the qualitative results in the landmark localization task as shown in Figure \ref{fig:qualitative_SOTA}.

Localization performance results across correctly classified teeth can be seen at Table \ref{tab:SOTA_MEDE}. In terms of MEDE, CHaRNet beats TSMDL and ALIIOS across all dentition types, reducing the micro-averaged error to 0.80 mm. More importantly, CHaRNet achieves a macro-averaged error of 1.12 mm versus 1.90 mm and 2.17 mm from TSMDL and ALIIOS, respectively.  When looking at MSR, we see that CHaRNet surpasses TSMDL and ALIIOS in all dentition types but in 00 and 10 subclasses, where TSMDL achieves slightly better results. These results show CHaRNet improves performance across complex dentitions with irregular geometries and missing teeth, which are understudied in previous works. This is exemplified in Figure \ref{fig:qualitative_SOTA}: on standard dentitions (a) where the segmentation predictions are correct, ALIIOS and, especially, TSMDL deliver great results in landmark localization. However, in complex dentition cases (b), the segmentation usually fails and propagates this error to the individual tooth landmark detection models of ALIIOS and TSMDL. This failed segmentation does not allow a subsequent landmarks placement, since the individual tooth meshes are mostly incorrect. As can be seen, CHaRNet tends to behave better across standard and complex dentitions.

Table \ref{tab:execution_times} compares the three methods in terms of time efficiency in CPU and GPU. CHaRNet is $4.5\times$ times faster in CPU and $14.8\times$ on GPU than TSMDL. ALIIOS is orders slower in both, especially in CPU. These big differences can be explained not only by the segmentation stage being time-consuming, but also by the necessity of TSMDL and ALIIOS to perform individual inferences for landmark localization for every tooth detected at the segmentation stage. ALIIOS is especially slower at individual tooth landmark localization given that five inferences per tooth must be performed. 

In conclusion, CHaRNet achieves the best (i) tooth classification, (ii) landmark localization, and (iii)  time efficiency at inference, allowing robust real-time landmark localization across any kind of dentition.

\begin{table*}[t]
\centering
\caption{Tooth/landmarks presence classification performance: F1-scores for ALIIOS, TSMDL, and CHaRNet. Results are shown for each dentition type, micro and macro-averaged. \textbf{Bold} indicates the best result for each dentition subclass.}
\renewcommand{\arraystretch}{1.18}
\resizebox{\textwidth}{!}{ 
\begin{tabular}{lrrrrrrrrrrrr}
\toprule
Model & 00 & 01 & 02 & 03 & 04 & 10 & 11 & 12 & 13 & 14 & Macro-avg & Micro-avg \\

\midrule
ALIIOS  & 0.99 & 0.97 & 0.93 & 0.83 & 0.91 & 0.99 & 0.93 & 0.91 & 0.85 & 0.62 & 0.89 & 0.98 \\
TSMDL  & 0.99 & 0.97 & 0.93 & 0.83 & 0.91 & 0.99 & 0.93 & 0.91 & 0.85 & 0.62 & 0.89 & 0.98 \\
CHaRNet & 0.99 & \textbf{0.98} & \textbf{0.95} & \textbf{0.85}  & \textbf{0.93} & 0.99 & \textbf{0.98} & \textbf{0.92} & \textbf{0.90} & \textbf{0.68} & \textbf{0.92} &  \textbf{0.99} \\
\bottomrule
\end{tabular}
}
\label{tab:SOTA_f1}
\end{table*}

\begin{table*}[t]
\centering
\caption{Landmarks localization performance: MEDE (\textdownarrow) and MSR (\textuparrow) at 1 mm for ALIIOS, TSMDL, and CHaRNet. Results are shown for each dentition type, micro and macro-averaged. \textbf{Bold} indicates the best result for each dentition subclass.}
\renewcommand{\arraystretch}{1.18}
\resizebox{\textwidth}{!}{ 
\begin{tabular}{lrrrrrrrrrrrr}
\toprule
Model & 00 & 01 & 02 & 03 & 04 & 10 & 11 & 12 & 13 & 14 & Macro-avg & Micro-avg \\
\midrule
\textit{MEDE} (\textdownarrow) &  &  &  &  &  &  &  &  &  &  &  & \\
\midrule
ALIIOS  & 0.93 & 1.49 & 1.64 & 3.15 & 4.39 & 0.94 & 1.76 & 1.56 & 2.69 & 3.10 & 2.17 & 1.12 \\
TSMDL   & 0.59 & 1.26 & 1.30 & 2.75 & 2.85 & \textbf{0.64} & 1.48 & 1.51 & 2.31 & 3.27 & 1.80 & 0.83 \\
CHaRNet & \textbf{0.56} & \textbf{0.76} & \textbf{0.99} & \textbf{1.66} & \textbf{2.14} & \textbf{0.64} & \textbf{0.98} & \textbf{0.67} & \textbf{0.85} & \textbf{1.91} & \textbf{1.12} & \textbf{0.70} \\

\midrule
\textit{MSR} (\textuparrow) &  &  &  &  &  &  &  &  &  &  &  & \\
\midrule
ALIIOS  & 70.2 & 55.9 & 55.0 & 22.7 & 12.5 & 70.5 & 53.9 & 58.2 & 41.1 & 40.0 & 48.0 & 65.7 \\
TSMDL  & \textbf{87.6} & 71.3 & 70.3 & 36.4 & 15.5 & \textbf{85.1} & 67.3 & 65.9 & 52.7 & 37.7 & 59.0 & 79.4 \\
CHaRNet & 85.2 & \textbf{77.6} & \textbf{79.5} & \textbf{49.1} & \textbf{29.5} & 79.7 & \textbf{70.9} & \textbf{80.1} & \textbf{69.0} & \textbf{63.8} & \textbf{68.5} &  \textbf{79.8} \\
\bottomrule
\end{tabular}
}
\label{tab:SOTA_MEDE}
\end{table*}

\begin{table}[t]
\centering
\caption{ Inference time performance: Mean execution time and standard deviation (in seconds) over the test set for ALIIOS, TSMDL, and CHaRNet. \textbf{Bold} denotes the best result for each device.}
\renewcommand{\arraystretch}{1.0}
\begin{tabular}{lrr}
\toprule
Model & CPU (s) & GPU (s) \\
\midrule
ALIIOS  & 612.13 $\pm$ 24.27 & 24.70 $\pm$ 3.54 \\
TSMDL   & 3.89 $\pm$ 0.85    & 3.55 $\pm$ 0.81 \\
CHaRNet & \textbf{0.86 $\pm$ 0.06} & \textbf{0.24 $\pm$ 0.03} \\
\bottomrule
\end{tabular}
\label{tab:execution_times}
\end{table}

\section{Discussion} \label{sec:discussion}

The results demonstrate the effectiveness of the proposed CHaRM and CHaR module in improving 3D landmark detection for point cloud-based models, particularly in challenging scenarios involving incomplete dental models. Across multiple architectures, CHaR-based models consistently outperform their baseline counterparts on both MEDE and MSR, underscoring the impact of enhancing the encoding of local geometric information. This improvement is especially evident in cases with missing teeth, where baseline models often fail to capture the subtle cues necessary for accurate landmark localization. When comparing performance across different dentition types, all models perform better on standard dental models than on incomplete ones. For instance, CHaRNet (CHaR-based PointMLP) achieves a MEDE of 0.56 mm and an MSR of 85.2\% on standard models, highlighting its ability to handle simpler cases with precision. However, macro-averaged results that include more complex scenarios exhibit a higher error (1.12 mm MEDE) and lower success ratio (68.5\% MSR), suggesting that the absence of certain teeth disrupts key structural patterns used for landmark detection. Although the CHaR module significantly narrows this performance gap, there remains a clear need for methods better equipped to handle incomplete geometries. Qualitative analyses reinforce the superiority of CHaR-based models. Visual inspections, as shown in Figure~\ref{fig:failures}, reveal that while baseline models frequently misplace landmarks in regions with missing teeth, CHaR-based models are more robust, even when direct geometric cues are limited. The inferred landmark placements in these complex regions highlight the CHaR module’s capability to reason about plausible geometries in the face of missing data.


When compared with SOTA methods, TSMDL and ALIIOS, CHaRNet delivers significant improvements in three key aspects. (i) Better teeth classification is achieved, which is critical for visually acceptable results in complex dentition where segmentation errors could catastrophically hurt individual tooth landmarks placement, as shown in Figure \ref{fig:complex_dent}. (ii) Better landmark localization is achieved, mainly because of the superiority of PointMLP as point cloud feature extractor and the refinement of heatmaps' outputs with CHaR module. (iii) Far better inference time efficiency, reducing $711\times$ and $4.5\times$ times ALIIOS and TSMDL latency, respectively, in CPU, and $102\times$ and $14.8\times$ times ALIIOS and TSMDL latency, respectively, in GPU. In absolute terms, 0.86 and 0.24 seconds in CPU and GPU, respectively, allow CHaRNet to be used in real-time clinical dental applications or processes.

Despite these promising results, certain limitations suggest directions for future work. Although the CHaR module significantly enhances performance in complex cases, current MEDE and MSR values leave room for improvement. Approaches that combine the CHaR module with additional processing steps or other network modules could be explored to address more severe or irregular forms of dental model incompleteness. Furthermore, because the CHaR module appears to rely on specific architectural features of PointMLP, efforts to generalize it to a broader range of network designs could expand its utility. Finally, while the focus here is on landmark detection accuracy, real-time processing and scalability to larger point clouds and dataset sizes remain open questions that merit further investigation.

\section{Conclusion}  \label{sec:conclusion}

This work introduces CHaRM, a novel approach to enhance 3D landmark detection in point cloud-based models, particularly for dental applications. The results demonstrate that the CHaR module significantly improves performance across multiple point cloud architectures, as evidenced by consistent reductions in MEDE and increases in MSR for both standard dental models and cases involving missing teeth. By effectively grouping landmarks associated with each tooth, the CHaR module enables the model to adapt dynamically based on the tooth's presence or absence, assigning landmarks to a null point if the tooth is missing or accurately localizing them within the dentition if the tooth is present. This capability addresses a key challenge in dental landmark detection, where structural irregularities and missing teeth often hinder model performance.

The superior performance of the CHaR-based version in every model, particularly in PointMLP, highlights its suitability for incorporating the proposed module, achieving SOTA results with a MEDE of 0.56 mm in standard dental models and a robust performance of 1.12 mm across all dentition types. Visual findings also validate the efficacy of the CHaR module in improving the accuracy and robustness of landmark detection, particularly in challenging scenarios. These SOTA results are achieved by eliminating the segmentation step typically found in the literature, which allowed a drastic inference time reduction, making CHaRNet suitable for real-time applications.

To contribute to the advancement of research in this domain, we are making the dataset (IOSLandmarks-1k) and the code used in this study publicly available. This effort aims to address the lack of "open spirit" that is prevalent in fields such as orthodontics, in contrast to the artificial intelligence community, where openness and collaboration are common. We hope that this initiative will encourage others in the orthodontic community to share their datasets, allowing fair comparisons between AI methods, attracting more practitioners to the field of computer-assisted orthodontics, and fostering the development of new domain-specific technologies.

Future work will explore ways to further enhance the performance of the module in complex cases, generalize its application to a wider range of architectures, and evaluate its scalability and efficiency in larger datasets or real-time applications. By addressing these challenges, this research aims to contribute to the development of more effective and versatile 3D landmark detection systems.


\bibliographystyle{unsrt}  
\bibliography{references}

\end{document}